\documentclass{article}

\usepackage[main, final]{neurips_2026}
\usepackage{amsmath}
\usepackage{amssymb}
\usepackage{mathtools}
\usepackage{longtable}
\usepackage{caption}
\usepackage{graphicx}
\usepackage{multirow}
\usepackage{makecell}
\usepackage[utf8]{inputenc} % allow utf-8 input
\usepackage[T1]{fontenc}    % use 8-bit T1 fonts
\usepackage{hyperref}       % hyperlinks
\usepackage{url}            % simple URL typesetting
\usepackage{booktabs}       % professional-quality tables
\usepackage{amsfonts}       % blackboard math symbols
\usepackage{nicefrac}       % compact symbols for 1/2, etc.
\usepackage{microtype}      % microtypography
\usepackage{xcolor}         % colors
\usepackage{tabularx}

\title{SPERA: Spherical Prior EEG Foundation Model with Geometry- and Frequency-Aware Latent Prediction}

\author{
  \textbf{Minsu Kim}\textsuperscript{1},
  \textbf{Ye-Sung Kim}\textsuperscript{2},
  \textbf{Hyeseong Jeon}\textsuperscript{2}, \\
  \textbf{Wooseok Hyung}\textsuperscript{2},
  \textbf{Joshua Lee}\textsuperscript{2},
  \textbf{Chang-Hwan Im}\textsuperscript{3}\thanks{Corresponding author. \texttt{ich@hanyang.ac.kr}} \\
  \textnormal{\textsuperscript{1}Department of Electronic Engineering, Hanyang University, Seoul, Republic of Korea} \\
  \textnormal{\textsuperscript{2}Department of Artificial Intelligence, Hanyang University, Seoul, Republic of Korea} \\
  \textnormal{\textsuperscript{3}Department of Biomedical Engineering, Hanyang University, Seoul, Republic of Korea}
}

\begin{document}

\maketitle

\begin{abstract}
    Electroencephalography (EEG) provides a non-invasive measure of ongoing neural activity, but building general-purpose EEG models remains challenging due to the heterogeneity of subjects, devices, and electrode montages. 
    Existing EEG foundation models predominantly rely on reconstruction-based objectives defined on the observed signal, which contains both neural and non-neural components.
    We introduce the Spherical Prior EEG Representation Architecture (SPERA{\setcounter{footnote}{0}\renewcommand{\thefootnote}{\fnsymbol{footnote}}%
    \footnote[2]{Code: \url{https://github.com/sees4321/SPERA_EEG}}}), an EEG foundation model that adopts the joint-embedding predictive architecture (JEPA) to predict in latent space.
    SPERA introduces a Legendre-polynomial spatial prior, incorporated into attention to encode varying scalp electrode geometries. Two further components adapt the model to EEG: factorized temporal and spatial attention interleaved with periodic full-attention blocks, and a relational spectral regularizer aligning latent similarity structure with spectral views.
    Pretrained on approximately 80,000 hours of EEG from 29,048 subjects across 106 datasets, SPERA achieves the highest average balanced accuracy across nine downstream tasks spanning clinical, cognitive, and BCI applications. 
    SPERA further exhibits strong parameter efficiency under linear probing and robustness across varying recording conditions, suggesting its potential as a general-purpose backbone for diverse EEG analyses.
\end{abstract}

\section{Introduction}
\label{sec:intro}
Electroencephalography (EEG) reflects dynamic neural activity and underlying brain states.
With high temporal resolution and ease of acquisition, EEG is widely used across clinical diagnosis, cognitive neuroscience, and brain-computer interface (BCI) applications~\cite{smith2005eeg, motamedi2014signal, bell2012using, varbu2022past}.
Deep learning models have achieved strong performance in interpreting these signals~\cite{altaheri2023deep, saha2018classification}.

Despite this progress, building a general-purpose EEG model remains challenging due to the heterogeneity of EEG data.
EEG signals vary significantly across individuals~\cite{saha2020intra} and are recorded under diverse experimental conditions with different devices and electrode configurations~\cite{bigdely2015prep}.
This makes it difficult to learn representations that transfer across datasets and subjects.

Inspired by foundation models in vision and language~\cite{ravi2024sam, grattafiori2024llama}, recent EEG foundation models such as LaBraM~\cite{jiang2024large}, CBraMod~\cite{wang2024cbramod}, and REVE~\cite{ouahidi2025reve} have explored scalable self-supervised pretraining, using masked reconstruction or token-prediction objectives whose targets are derived from the input signal.
EEG recordings contain both neural activity and non-neural components such as muscle and ocular artifacts. 
Reconstruction-based objectives define their prediction targets on all components present in the observed signal, which motivates considering prediction targets defined at the representation level. 
We therefore adopt the joint-embedding predictive architecture (JEPA)~\cite{assran2023self}.

We propose the Spherical Prior EEG Representation Architecture (SPERA), a representation learning framework for EEG built on latent-space prediction.
Beyond this objective, our central contribution is a Legendre polynomial spherical prior incorporated into attention, providing a learnable pairwise geometric bias that reflects the spherical structure of scalp EEG and applies to arbitrary montages. 
Two further components adapt the backbone to EEG: factorized temporal and spatial attention interleaved with periodic full-attention blocks for cross-axis interactions, and a relational spectral regularizer that induces frequency-aware latents by aligning pairwise similarity structure with spectral views.

Building on these components, we pretrain SPERA on a multi-domain EEG corpus comprising approximately 80,000 hours of recordings from 29,048 subjects across 106 datasets.
We evaluate SPERA on nine diverse downstream tasks spanning clinical, cognitive, and BCI applications and compare it with prior EEG foundation models. We further investigate the scaling behavior of latent-space prediction under EEG-aware design, with a focus on parameter efficiency under linear probing (LP).

\section{Related work}
\label{sec:related_work}

\paragraph{Self-supervised objectives for EEG}
Most existing EEG foundation models~\cite{jiang2024large, wang2024cbramod, ouahidi2025reve, ma2026codebrain, li2025gram} rely on masked modeling objectives with targets derived from the input signal, including reconstructed signal patches and discrete neural tokens. 
Such targets therefore include both neural and non-neural components of the observed signal.
JEPA models~\cite{assran2023self, assran2025v} instead predict in latent space, and several EEG models have moved in this direction: latent prediction against a momentum-updated target encoder~\cite{mohammadi2024eeg2rep}, representation alignment paired with auxiliary patch reconstruction~\cite{wang2024eegpt}, generative diffusion whose denoising activations serve as representations~\cite{puah2025eegdm}, and direct applications of JEPA to EEG~\cite{guetschel2024s, hojjati2025video, panchavati2026laya}. 
However, none of these explicitly encode the pairwise geometric relationships between electrodes, which vary across montage configurations.
We use a latent prediction objective over a montage-variable token grid, with masking of temporal regions and spatial neighborhoods defined by 3D electrode coordinates, and scale it to a multi-domain corpus spanning over a hundred datasets.

\paragraph{Electrode geometry in EEG modeling}
A central challenge for EEG foundation models is the heterogeneity of electrode montages across datasets.
BIOT adds channel embeddings and relative position embeddings~\cite{yang2023biot}, CBraMod uses asymmetric conditional positional encoding to adapt to diverse EEG formats~\cite{wang2024cbramod}, and REVE introduces a 4D positional encoding scheme for arbitrary electrode arrangements~\cite{ouahidi2025reve}.
Recent JEPA-style EEG models also introduce spatial inductive biases, including 3D coordinate-based positional embeddings~\cite{guetschel2024s}, video-like spatio-temporal tokenization~\cite{hojjati2025video}, and coordinate-aware channel mixing with Fourier features~\cite{panchavati2026laya}.
We instead encode electrode geometry directly into attention using a Legendre-based spherical prior, providing a learnable pairwise bias derived from the biophysical geometry of the scalp~\cite{perrin1989spherical,carvalhaes2015surface}.

\paragraph{Spatio-temporal modeling in EEG}
EEG signals exhibit distinct spatial and temporal dependencies, motivating attention factorization along each axis.
A related factorized design has been used in video understanding, where TimeSformer and ViViT apply temporal and spatial attention sequentially within each block~\cite{bertasius2021space,arnab2021vivit}. 
In EEG, CBraMod models spatial and temporal dependencies in parallel using criss-cross attention~\cite{wang2024cbramod}, while CSBrain~\cite{zhou2026csbrain} combines cross-scale spatio-temporal tokenization with structured sparse attention over local temporal windows and anatomical regions.
State-space models use Mamba-based backbones for EEG sequence modeling~\cite{wang2025eegmamba}. LuMamba uses learned-query cross-attention to unify variable electrode configurations before temporal modeling~\cite{broustail2026lumamba}.
These approaches capture spatial and temporal structure in different ways.
However, our setting introduces two additional design considerations: EEG channel montages vary across datasets, unlike the fixed spatial grids used in these video models, and purely factorized attention restricts direct cross-axis interaction.
We address these considerations by combining explicit electrode-geometry encoding with a hybrid backbone that interleaves factorized and full-attention blocks.

\paragraph{Spectral objectives in EEG pretraining}
Frequency components are key features of EEG signals, motivating spectral objectives in several pretraining methods.
LaBraM trains its tokenizer with vector-quantized neural spectrum prediction~\cite{jiang2024large}, and Pradeepkumar et al.~\cite{pradeepkumar2025tokenizing} introduce time-frequency motif learning for frequency-aware EEG tokenization.
We complement this line of work with a relational spectral regularizer that aligns pairwise similarity structure between latents and spectral views, without predicting absolute spectral values.

\section{Methods}
\label{sec:methods}

\begin{figure}[t]
    \centering
    \includegraphics[width=1\textwidth]{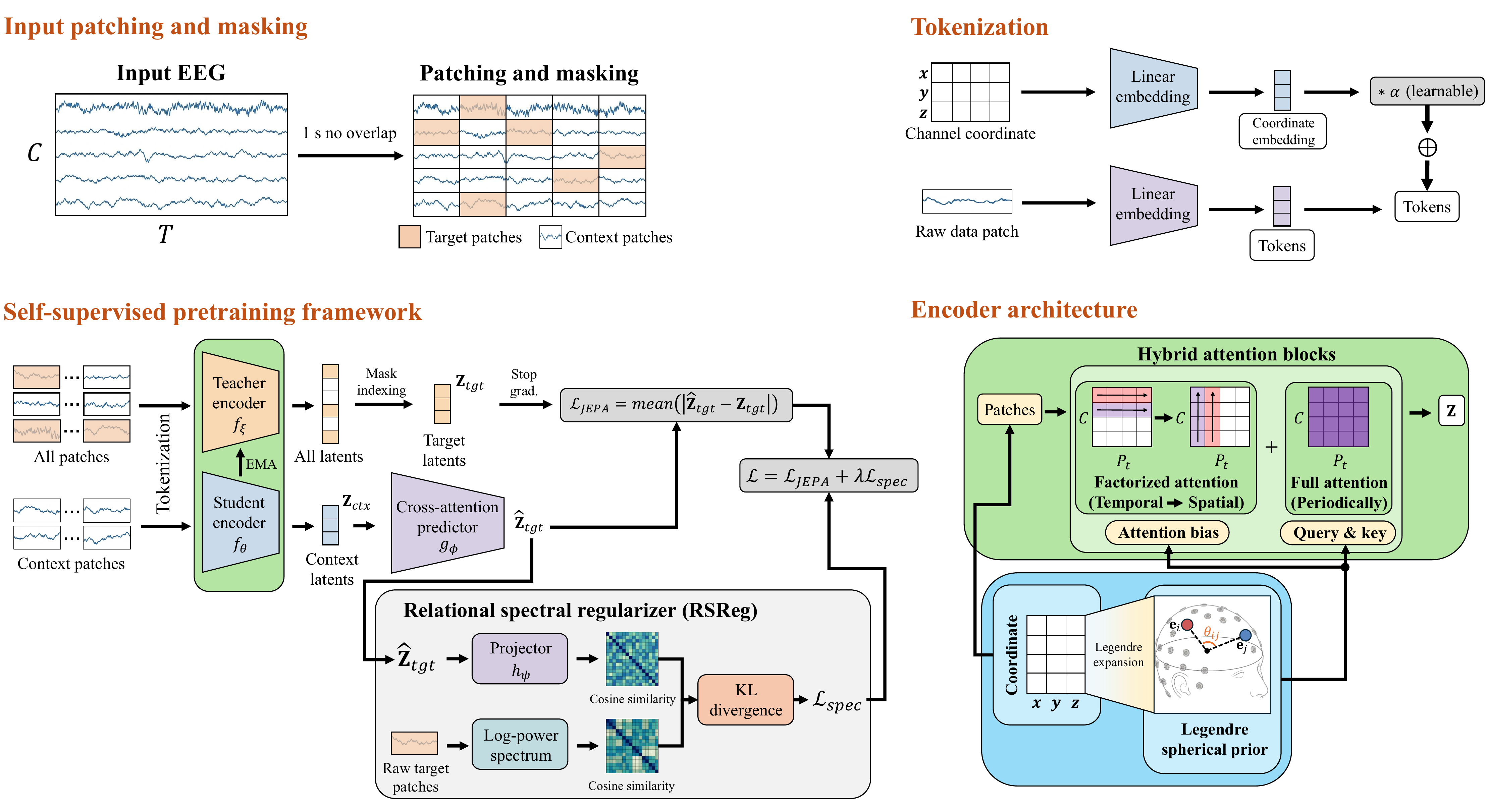}
    \caption{Overview of the SPERA pretraining framework.}
    \label{fig:overview}
\end{figure}

We denote an EEG recording as $\mathbf{X} \in \mathbb{R}^{C \times T}$, where $C$ is the number of electrodes and $T$ is the number of time samples.
Each electrode is associated with a 3D anatomical coordinate, collectively represented as $\mathbf{E} \in \mathbb{R}^{C \times 3}$.
To tokenize $\mathbf{X}$, each of the $C$ channels is divided into $P_t$ non-overlapping 1-second temporal patches at a sampling rate of $f_s =$ 200 Hz, resulting in a total of $N = C \cdot P_t$ tokens.
Each token is uniquely indexed by a (channel, time-patch) tuple $(c, p)$ on the spatio-temporal grid.
Unlike token sequences in language or vision, EEG signals have dataset-specific spatio-temporal grids, as channel montages vary across datasets. 
This cross-dataset variability and the two-axis structure of the token space motivate the architectural choices described in this section.

\subsection{Self-supervised objective: spatio-temporal JEPA}
\paragraph{JEPA objective}
Figure~\ref{fig:overview} illustrates the overall pretraining framework. 
Raw EEG contains muscle and ocular artifacts that persist after preprocessing, so an input-space reconstruction objective requires reproducing them alongside neural activity.
We adopt JEPA~\cite{assran2023self}, which predicts latent representations of masked target regions from visible context.
SPERA consists of a student encoder $f_\theta$ and a teacher encoder $f_\xi$; the teacher receives no gradients and is updated as an exponential moving average (EMA) of $f_\theta$.
On the spatio-temporal grid, we partition the token indices into disjoint context and target sets, $\mathcal{C}$ and $\mathcal{T}$.
The student encodes only the context tokens to produce $\mathbf{Z}_{\text{ctx}}$. 
The teacher encodes the full token sequence, and $\mathbf{Z}_{\text{tgt}}$ is obtained by selecting the teacher outputs at the target positions.
A two-layer cross-attention predictor $g_\phi$ predicts $\hat{\mathbf{Z}}_{\text{tgt}}$ from $\mathbf{Z}_{\text{ctx}}$.
For each target position, a query is formed by adding the target coordinate embedding to a single learnable vector shared across all target positions, and is refined by cross-attending to $\mathbf{Z}_{\text{ctx}}$.
Following prior work~\cite{assran2025v}, the model is trained to minimize the $\ell_1$ distance in latent space:
\begin{equation}
    \mathcal{L}_{\text{JEPA}} = \frac{1}{|\mathcal{T}|D} \sum_{(c,p) \in \mathcal{T}} \left\| \hat{\mathbf{z}}_{c,p} - \mathbf{z}_{c,p} \right\|_1,
\end{equation}
where $\hat{\mathbf{z}}_{c,p}$ and $\mathbf{z}_{c,p}$ denote the predicted and teacher latents at position $(c,p)$, and $D$ is the embedding dimension.

\paragraph{Masking strategy}
For each sample, temporal masking is activated with probability 0.8, and spatial masking is activated independently with probability 0.8.
If neither is activated, we default to temporal masking.
Temporal masks follow a semantic subsequence preserving (SSP) strategy~\cite{mohammadi2024eeg2rep}: rather than masking random individual time patches, we preserve contiguous blocks of patches to ensure the context carries coherent temporal structure. 
Spatial masks select a random electrode as a center and mask its $k$-nearest neighbors based on 3D Euclidean distance, with $k = \max(1, \lfloor \rho_s \cdot C \rceil)$ for a sampled spatial ratio $\rho_s$. 
Masking ratios are drawn uniformly from [0.15, 0.35] for time and [0.10, 0.30] for space ($\rho_s$).

\paragraph{Student augmentation}
We apply EEG-specific perturbations to the student's context input: temporal amplitude scaling and additive Gaussian noise, together with Gaussian jitter to electrode coordinates.
Augmentation hyperparameters are reported in Appendix~\ref{app:impl}.

\subsection{Encoder architecture}
Dense full attention allows interactions among all spatio-temporal token pairs but does not explicitly distinguish temporal and spatial dependencies.
We adopt a hybrid backbone of factorized attention blocks interleaved with periodic full-attention blocks for cross-axis interactions.

\paragraph{Hybrid attention blocks}
The base block factorizes attention along each axis separately. 
A \textit{temporal attention} pass operates independently over the $P_t$ time patches within each channel, followed by a \textit{spatial attention} pass over the $C$ channels at each time patch. 
Every fourth block is a full self-attention block that attends jointly over all $N = C \cdot P_t$ tokens, enabling cross-axis interactions. 
Both attention block types share standard transformer design choices: a pre-norm architecture with layer normalization (LN), a SwiGLU MLP~\cite{shazeer2020glu}, and rotary position embeddings (RoPE)~\cite{su2024roformer} applied along the temporal axis (in both the encoder and the predictor). 

\subsection{Electrode geometry encoding with spherical prior}
\label{sec:spatial}
To handle varying electrode montages across datasets, we incorporate simple coordinate embeddings and Legendre-based spherical priors.
The Legendre prior is implemented in two forms: a pairwise Legendre bias for factorized attention blocks and Legendre anchor encoding for full-attention blocks. 
The two are assigned to attention types based on computational considerations analyzed in Section~\ref{sec:ablation}.

\paragraph{Coordinate embedding}
A direct approach to spatial conditioning is a per-token additive embedding derived from electrode coordinates.
We normalize each coordinate onto the unit sphere and pass it through a linear projection. 
The resulting channel-wise embedding is scaled by a learnable scalar $\alpha$, initialized to zero, and shared across all temporal patches of the same electrode.

\paragraph{Legendre spatial bias}
Scalp EEG electrodes lie approximately on a spherical head surface, and we exploit this structure to build a biophysically motivated spatial prior on the sphere itself.
Legendre polynomials provide a natural basis for parameterizing pairwise functions of angular separation on the sphere and have been used in conventional EEG analysis~\cite{perrin1989spherical, carvalhaes2015surface}. 
We therefore use a truncated Legendre expansion to encode pairwise electrode geometry as an additive attention bias.

Within each factorized attention block, the spatial attention pass computes attention over $C$ channels at a fixed time patch. To encode pairwise geometric relationships between electrodes, we introduce a learnable additive attention bias $b_{ij}$ between channel $i$ and channel $j$, defined as a truncated Legendre polynomial expansion of their cosine similarity on the unit sphere:
\begin{equation}
    b_{ij} = \sum_{\ell=0}^{L} w_\ell \, P_\ell\!\left(\mathbf{e}_i^\top \mathbf{e}_j\right),
\end{equation}
where $\mathbf{e}$ denotes unit-normalized coordinates, $L$ is the maximum polynomial degree, $P_\ell$ is the $\ell$-th Legendre polynomial, and $w_\ell$ are learnable scalar coefficients. 
Unlike classical formulations that fix $w_\ell$ to a closed-form decay (e.g., $(2\ell+1)/[\ell^m(\ell+1)^m]$ in spherical splines), we let the coefficients be learned end-to-end.
This bias is pairwise, geometry-aware, and dataset-agnostic, requiring no assumption about the number or identity of electrodes.

\paragraph{Legendre anchor encoding}
The pairwise Legendre bias in Eq.~(2) directly encodes the angular relationship between each pair of electrodes. In the dense implementation, applying this pairwise bias in full-attention blocks requires materializing an $N\times N$ score-bias matrix, which incurs substantial memory and computational overhead. 
We therefore construct an anchor-based alternative from the same Legendre geometry. 
Instead of expanding the relationship between every pair of electrodes, we represent each electrode by its Legendre-expanded similarities to a finite set of learnable anchor directions on the unit sphere.
These anchor-based geometry encodings are then injected into the queries and keys of the full-attention blocks.

We compute a per-channel spatial feature $\boldsymbol{\phi}_c$ using a set of $M$ learnable anchor points $\hat{\mathbf a}_m$ on the unit sphere, expanding the cosine similarity between each electrode and each anchor via Legendre polynomials:
\begin{equation}
g_{c,m}=\sum_{\ell=0}^{L}\omega_{m,\ell}\,
P_{\ell}\!\left(\mathbf{e}_c^{\top}\hat{\mathbf a}_m\right),
\qquad
\boldsymbol{\phi}_c=\mathrm{LN}\!\left(W_{\mathrm{proj}}\,[g_{c,1},\ldots,g_{c,M}]\right),
\end{equation}
\begin{equation}
\tilde{\mathbf q}_i=\mathbf q_i+\Gamma_Q \boldsymbol{\phi}_{c_i},
\qquad
\tilde{\mathbf k}_j=\mathbf k_j+\Gamma_K\boldsymbol{\phi}_{c_j}, 
\qquad
s_{ij}=\frac{\tilde{\mathbf q}_i^{\top}\tilde{\mathbf k}_j}{\sqrt d},
\end{equation}
where $\omega_{m,\ell}$ are learnable anchor-specific coefficients and $W_{\mathrm{proj}}$ is a linear projection.
Here, $c_{i}$ denotes the channel index of token $i$, and $d$ is the attention head dimension. 
The geometry feature $\boldsymbol{\phi}_c$ is added to queries and keys via projections $\Gamma_Q$ and $\Gamma_K$, applied after temporal RoPE.
Expanding Eq.~(4) gives four terms: content--content, an anchor-induced geometry--geometry term, and two content--geometry cross terms (see Appendix~\ref{app:prior:decomp} for details).
The coefficients in Eqs.~(2)--(3), as well as the anchor directions $\hat{\mathbf a}_m$, are shared globally across layers and heads.
In the anchor-based formulation, layer-specific adaptation is introduced through the block-specific projections $\Gamma_Q, \Gamma_K$.

\subsection{Relational spectral regularizer}
Frequency components are key features of EEG signals, but incorporating spectral objectives into a JEPA framework is not straightforward.
Direct spectrum reconstruction would introduce an absolute signal-derived target, whereas our goal is to regularize spectral relationships without reconstructing spectral values.
Therefore, we introduce a relational spectral regularizer (RSReg) that aligns the pairwise similarity structure of the predicted representations with that of the corresponding spectral targets.
This design is inspired by relation-preserving distillation~\cite{passalis2018learning, park2019relational}, which supervises pairwise relational structure.
Here, the target relations are computed deterministically from spectral views of EEG patches, not from another network.

For each target token $(c, p)$, $\mathbf{s}_{c,p}$ is the log-power spectrum of its 1-second raw patch, computed via a Hann-windowed FFT using frequency bins within the 1--45 Hz band, and z-score normalized across frequency bins.
The predictor output $\hat{\mathbf{z}}_{c,p}$ is passed through a small projector $h_\psi$ to obtain $\mathbf{r}_{c,p} = h_\psi(\hat{\mathbf{z}}_{c,p})$. 
We align the pairwise similarity matrices of $\mathbf{r}$ and $\mathbf{s}$ through a KL divergence loss:
\begin{equation}
    \mathcal{L}_{\text{spec}} = \mathrm{KL}\!\left(\sigma\!\left(\mathbf{S}\mathbf{S}^\top / \tau_s\right) \,\Big\|\, \sigma\!\left(\mathbf{R}\mathbf{R}^\top / \tau_z\right)\right),
\end{equation}
where $\mathbf{S}$ and $\mathbf{R}$ are row-wise $\ell_2$-normalized matrices of $\mathbf{s}$ and $\mathbf{r}$, respectively, $\sigma(\cdot)$ denotes row-wise softmax, and $\tau_s, \tau_z$ are temperature hyperparameters. 
Similarity matrices are computed per recording, with diagonal entries masked before softmax.
The KL loss is averaged over target-token rows within each recording, then equally across recordings in the effective batch.
This preserves relative spectral relationships among tokens without constraining their absolute spectral values, inducing frequency-aware latent representations through relational spectral supervision.
The total training loss is:
\begin{equation} 
    \mathcal{L} = \mathcal{L}_{\text{JEPA}} + \lambda \, \mathcal{L}_{\text{spec}}.
\end{equation}

\section{Experiments}
\label{sec:experiments}

\subsection{Pretraining}
\label{sec:pretraining}
\paragraph{Dataset collection}
We collected a large-scale pretraining corpus spanning approximately 80,000 hours of EEG recordings from 29,048 subjects, yielding 2.67 million channel-hours and 9.60 billion tokens in total. 
To the best of our knowledge, this is the largest multi-domain EEG pretraining corpus reported for a general-purpose EEG foundation model. 
The corpus is organized into three categories (clinical, cognitive, and BCI), comprising 82.08\%, 16.21\%, and 1.71\% of the total recording time, respectively.
The pretraining dataset exhibits substantial heterogeneity in electrode montages, covering a wide range of channel counts and layouts across sources. 
A summary of the dataset composition and a complete list of included sources are provided in Appendix~\ref{app:ptds}. 

\paragraph{Preprocessing pipeline}
We apply a standardized preprocessing pipeline to all recordings. 
Power line noise is removed via a notch filter at 50 or 60 Hz depending on the recording country. 
EEG recordings are then bandpass filtered between 0.5 and 75 Hz and resampled to $f_s =$ 200 Hz. 
Each channel is $z$-score normalized to account for cross-dataset variability in electrode impedance and amplifier gain. 
For datasets that provide digitized 3D electrode locations, we use the recorded coordinates directly; for those that specify only channel labels (e.g., the 10--20 system), we assign coordinates from the corresponding standard montage in MNE-Python~\cite{gramfort2014mne}.
In addition, amplitude values exceeding $\pm$ 15 standard deviations are clipped to suppress residual artifacts, as in~\cite{ouahidi2025reve}. 
Each recording is segmented into non-overlapping windows of either 10 or 60 seconds depending on the source task. 
Among the 60-second segments, those whose token count exceeds 4,096 are further split into 30-second chunks for training efficiency, preserving contiguous temporal structure.

\paragraph{Implementation details}
We train all models with AdamW~\cite{loshchilov2019decoupled}. 
The learning rate follows a cosine schedule with linear warmup, and the teacher encoder is updated via an EMA with a cosine momentum schedule. 
We pretrain four model sizes: SPERA-Tiny (18M), SPERA-Small (48M), SPERA-Base (194M), and SPERA-Large (396M). 
Full hyperparameter details are provided in Appendix~\ref{app:impl}.

\subsection{Experimental setup}
\label{sec:eval_protocol}
\paragraph{Downstream task datasets}
We evaluate SPERA on nine datasets spanning diverse clinical, cognitive, and BCI tasks: BCIC2020-3~\cite{jeong20222020}, BCIC-IV-2a~\cite{tangermann2012review}, PhysioNet-MI~\cite{goldberger2000physiobank,schalk2004bci2000}, FACED~\cite{chen2023large}, mental arithmetic (MAT)~\cite{zyma2019electroencephalograms}, ISRUC~\cite{khalighi2016isruc}, Mumtaz~\cite{Mumtaz2016}, TUAB~\cite{obeid2016temple}, and TUEV~\cite{obeid2016temple}. 
These nine tasks are the intersection of the CBraMod and REVE evaluation suites.
We adopt the same train/validation/test splits and preprocessing pipeline as baseline EEG foundation models~\cite{yang2023biot, jiang2024large, wang2024cbramod, ouahidi2025reve}.
Table~\ref{tab:dataset_summary} summarizes these datasets. 

\begin{table}[t]
    \centering
    \caption{Summary of downstream task datasets.}
    % \vspace{2mm}
    \label{tab:dataset_summary}
    \resizebox{\textwidth}{!}{
    \begin{tabular}{lccccc}
        \toprule
        \textbf{Task} & \textbf{Dataset} & \textbf{\# Samples} & \textbf{\# Channels} & \textbf{Duration (s)} & \textbf{\# Classes} \\
        \midrule
        Imagined speech classification& BCIC2020-3 & 6,000 & 64 & 3 & 5 \\
        \multirow{2}{*}{Motor imagery classification} & BCIC-IV-2a & 5,184 & 22 & 4 & 4 \\
         & PhysioNet-MI & 9,837 & 64 & 4 & 4 \\
        Emotion recognition & FACED & 10,332 & 32 & 10 & 9 \\
        Mental stress detection & MAT & 1,707 & 20 & 5 & 2 \\
        Sleep staging & ISRUC & 88,140 & 6 & 30 & 5 \\
        Mental disorder diagnosis & Mumtaz & 7,143 & 19 & 5 & 2 \\
        EEG abnormality detection & TUAB & 409,455 & 16 & 10 & 2 \\
        Event type classification & TUEV & 112,491 & 16 & 5 & 6 \\
        \bottomrule
    \end{tabular}
    }
\end{table}

\paragraph{Baseline models and evaluation protocol}
We compare SPERA against both non-foundation and foundation model baselines.
Non-foundation baselines include EEGNet~\cite{lawhern2018eegnet}, EEGConformer~\cite{song2022eeg}, SPaRCNet~\cite{jing2023development}, ContraWR~\cite{yang2023self}, CNN-Transformer~\cite{peh2022transformer}, FFCL~\cite{li2022motor}, ST-Transformer~\cite{song2021transformer}, EEG2Rep~\cite{mohammadi2024eeg2rep}, and EEGDM~\cite{puah2025eegdm}, while foundation model baselines include BIOT~\cite{yang2023biot}, LaBraM~\cite{jiang2024large}, CBraMod~\cite{wang2024cbramod}, REVE~\cite{ouahidi2025reve}, CSBrain~\cite{zhou2026csbrain}, CodeBrain~\cite{ma2026codebrain}, EEGMamba~\cite{wang2025eegmamba}, GRAM~\cite{li2025gram}, LuMamba~\cite{broustail2026lumamba}, EEGPT~\cite{wang2024eegpt}, and Laya~\cite{panchavati2026laya}.

We evaluate SPERA under two protocols, LP and linear probing followed by fine-tuning (LPFT). 
LP freezes the pretrained encoder and trains a linear classification head on mean-pooled token features, while LPFT additionally fine-tunes the encoder after the LP stage.
For each task, hyperparameters and checkpoints are selected on the validation set, and the held-out test set is used exclusively for final evaluation. 
Baseline results are extracted from the original publications~\cite{wang2024cbramod,ouahidi2025reve,zhou2026csbrain}; their downstream adaptation protocols vary across studies (e.g., CBraMod uses an MLP classification head).
Balanced accuracy is used as the primary metric; each experiment is repeated over five random seeds, and results are reported as mean $\pm$ standard deviation. 
For SPERA, the overall average is computed per seed as the unweighted mean across the nine datasets, and the reported standard deviation is taken across the resulting five seed-wise averages. Baseline results, including reported averages and standard deviations, are taken from the original publications.

\begin{table}[t]
    \centering
    \captionsetup{font=normalsize}
    \caption{Experimental results on nine downstream tasks. SPERA models are evaluated under the LPFT protocol, while baseline results are taken from the original publications using their reported downstream adaptation protocols. The results are reported as mean balanced accuracy $\pm$ standard deviation over five random seeds. The best performance for each task is highlighted in bold.}
    % \vspace{2mm}
    \label{tab:main}
    \resizebox{\textwidth}{!}{
    \begin{tabular}{lccccc}
        \toprule
        \textbf{Methods} & \textbf{BCIC2020-3} & \textbf{BCIC-IV-2a} & \textbf{PhysioNet-MI} & \textbf{FACED} & \textbf{MAT} \\
        \midrule
        EEGNet & $0.4413 \pm 0.0096$ & $0.4482 \pm 0.0094$ & $0.5814 \pm 0.0125$ & $0.4090 \pm 0.0122$ & $0.6770 \pm 0.0116$ \\
        EEGConformer & $0.4506 \pm 0.0133$ & $0.4696 \pm 0.0106$ & $0.6049 \pm 0.0104$ & $0.4559 \pm 0.0125$ & $0.6805 \pm 0.0123$ \\
        SPaRCNet & $0.4426 \pm 0.0156$ & $0.4635 \pm 0.0117$ & $0.5932 \pm 0.0152$ & $0.4673 \pm 0.0155$ & $0.6879 \pm 0.0107$ \\
        ContraWR & $0.4257 \pm 0.0162$ & $0.4678 \pm 0.0125$ & $0.5892 \pm 0.0133$ & $0.4887 \pm 0.0078$ & $0.6631 \pm 0.0097$ \\
        CNN-Transformer & $0.4533 \pm 0.0092$ & $0.4600 \pm 0.0108$ & $0.6053 \pm 0.0118$ & $0.4697 \pm 0.0132$ & $0.6779 \pm 0.0268$ \\
        FFCL & $0.4678 \pm 0.0197$ & $0.4470 \pm 0.0143$ & $0.5726 \pm 0.0092$ & $0.4673 \pm 0.0158$ & $0.6798 \pm 0.0142$ \\
        ST-Transformer  & $0.4126 \pm 0.0122$ & $0.4575 \pm 0.0145$ & $0.6035 \pm 0.0081$ & $0.4810 \pm 0.0079$ & $0.6631 \pm 0.0173$ \\
        \midrule
        BIOT  & $0.4920 \pm 0.0086$ & $0.4748 \pm 0.0093$ & $0.6153 \pm 0.0154$ & $0.5118 \pm 0.0118$ & $0.6875 \pm 0.0186$ \\
        LaBraM-Base  & $0.5060 \pm 0.0155$ & $0.4869 \pm 0.0085$ & $0.6173 \pm 0.0122$ & $0.5273 \pm 0.0107$ & $0.6909 \pm 0.0125$ \\
        CBraMod & $0.5373 \pm 0.0108$ & $0.5138 \pm 0.0066$ & $0.6417 \pm 0.0091$ & $0.5509 \pm 0.0089$ & $0.7256 \pm 0.0132$ \\
        CSBrain & $0.6004 \pm 0.0187$ & $0.5657 \pm 0.0071$ & $0.6304 \pm 0.0090$ & $0.5752 \pm 0.0042$ & $0.7558 \pm 0.0106$ \\
        REVE-Base & $0.5635 \pm 0.0123$ & $0.6396 \pm 0.0095$ & $0.6480 \pm 0.0140$ & $0.5646 \pm 0.0164$ & $0.7660 \pm 0.0355$ \\
        \midrule
        SPERA-Tiny & 0.4827 $\pm$ 0.0173 & 0.5326 $\pm$ 0.0138 & 0.6117 $\pm$ 0.0093 & 0.4849 $\pm$ 0.0155 & 0.6971 $\pm$ 0.0229 \\
        SPERA-Small & 0.5733 $\pm$ 0.0135 & 0.6085 $\pm$ 0.0156 & 0.6400 $\pm$ 0.0112 & 0.5132 $\pm$ 0.0167 & 0.7238 $\pm$ 0.0198 \\
        SPERA-Base & 0.6504 $\pm$ 0.0122 & 0.6816 $\pm$ 0.0147 & 0.6726 $\pm$ 0.0072 & 0.5754 $\pm$ 0.0175 & 0.7778 $\pm$ 0.0216 \\
        SPERA-Large & \textbf{0.6883 $\pm$ 0.0142} & \textbf{0.7023 $\pm$ 0.0144} & \textbf{0.6817 $\pm$ 0.0087} & \textbf{0.5919 $\pm$ 0.0186} & \textbf{0.8021 $\pm$ 0.0203} \\
        \midrule \midrule
        
        \textbf{Methods} & \textbf{ISRUC} & \textbf{Mumtaz} & \textbf{TUAB} & \textbf{TUEV} & \textbf{Average} \\
        \midrule
        EEGNet  & $0.7154 \pm 0.0121$ & $0.9232 \pm 0.0104$ & $0.7642 \pm 0.0036$ & $0.3876 \pm 0.0143$ & $0.5941 \pm 0.0037$ \\
        EEGConformer  & $0.7400 \pm 0.0133$ & $0.9308 \pm 0.0117$ & $0.7758 \pm 0.0049$ & $0.4074 \pm 0.0164$ & $0.6128 \pm 0.0044$ \\
        SPaRCNet  & $0.7487 \pm 0.0075$ & $0.9316 \pm 0.0095$ & $0.7896 \pm 0.0018$ & $0.4161 \pm 0.0262$ & $0.6156 \pm 0.0047$ \\
        ContraWR  & $0.7402 \pm 0.0126$ & $0.9195 \pm 0.0115$ & $0.7746 \pm 0.0041$ & $0.4384 \pm 0.0349$ & $0.6119 \pm 0.0053$ \\
        CNN-Transformer  & $0.7363 \pm 0.0087$ & $0.9305 \pm 0.0068$ & $0.7777 \pm 0.0022$ & $0.4087 \pm 0.0161$ & $0.6133 \pm 0.0045$ \\
        FFCL & $0.7277 \pm 0.0182$ & $0.9314 \pm 0.0038$ & $0.7848 \pm 0.0038$ & $0.3979 \pm 0.0104$ & $0.6085 \pm 0.0044$ \\
        ST-Transformer  & $0.7381 \pm 0.0205$ & $0.9135 \pm 0.0103$ & $0.7966 \pm 0.0023$ & $0.3984 \pm 0.0228$ & $0.6071 \pm 0.0048$ \\
        \midrule
        BIOT  & $0.7527 \pm 0.0121$ & $0.9358 \pm 0.0052$ & $0.7959 \pm 0.0057$ & $0.5281 \pm 0.0225$ & $0.6438 \pm 0.0044$ \\
        LaBraM-Base  & $0.7633 \pm 0.0102$ & $0.9409 \pm 0.0079$ & $0.8140 \pm 0.0019$ & $0.6409 \pm 0.0065$ & $0.6653 \pm 0.0031$ \\
        CBraMod & $0.7865 \pm 0.0110$ & $0.9560 \pm 0.0056$ & $0.8289 \pm 0.0022$ & $0.6671 \pm 0.0107$ & $0.6898 \pm 0.0051$ \\
        CSBrain & $0.7925 \pm 0.0030$ & $0.9643 \pm 0.0155$ & $0.8172 \pm 0.0043$ & $0.6903 \pm 0.0059$ & $0.7102$ \\
        REVE-Base & $0.7819 \pm 0.0078$ & 0.9644 $\pm$ 0.0097 & $0.8315 \pm 0.0014$ & $0.6759 \pm 0.0229$ & $0.7150 \pm 0.0057$ \\
        \midrule
        SPERA-Tiny & 0.7719 $\pm$ 0.0124 & 0.9236 $\pm$ 0.0034 & 0.8104 $\pm$ 0.0044 & 0.7035 $\pm$ 0.0273 & 0.6687 $\pm$ 0.0064 \\
        SPERA-Small & 0.7852 $\pm$ 0.0096 & 0.9450 $\pm$ 0.0048 & 0.8318 $\pm$ 0.0032 & 0.7384 $\pm$ 0.0258 & 0.7066 $\pm$ 0.0057 \\
        SPERA-Base & 0.8137 $\pm$ 0.0075 & 0.9637 $\pm$ 0.0018 & 0.8488 $\pm$ 0.0035 & 0.7684 $\pm$ 0.0283 & 0.7502 $\pm$ 0.0050 \\
        SPERA-Large & \textbf{0.8142 $\pm$ 0.0106} & \textbf{0.9702 $\pm$ 0.0016} & \textbf{0.8608 $\pm$ 0.0029} & \textbf{0.7869 $\pm$ 0.0286} & \textbf{0.7665 $\pm$ 0.0053} \\
        \bottomrule
    \end{tabular}
    }
\end{table}

\subsection{Downstream task results}
\label{sec:downstream_results}
Table~\ref{tab:main} summarizes LPFT performance across the nine downstream tasks and four SPERA model scales.
Per-task Welch tests over five seeds, Holm-corrected across the nine comparisons, find significant gains for SPERA-Base on five tasks: BCIC2020-3, TUEV, BCIC-IV-2a, ISRUC, and TUAB.
PhysioNet-MI is borderline ($p_{\mathrm{Holm}}=$ 0.052) and no task shows a significant loss.
Among the significant differences, the largest gains are observed on TUEV ($+$0.078; 6-class clinical event) and BCIC2020-3 ($+$0.050; 5-class imagined speech).
The differences on FACED, MAT, and Mumtaz did not reach statistical significance after Holm correction. 
Mumtaz in particular shows limited headroom, with balanced accuracy remaining generally high across foundation models and model scales.
SPERA performs consistently on both BCIC-IV-2a (22 channels) and PhysioNet-MI (64 channels), two motor-imagery tasks with different electrode layouts. 
Per-task differences with confidence intervals, for both SPERA-Base and SPERA-Large, are reported in Appendix~\ref{app:significance}.

\subsection{Model scaling analysis}
\label{sec:scaling}

Table~\ref{tab:main} shows that average downstream performance improves with model scale under LPFT. 
To further examine how the pretrained representations themselves scale, independently of full encoder fine-tuning, we evaluate the four model sizes (18M, 48M, 194M, and 396M) using LP across all nine downstream tasks. 
Per-task LP results are reported in Appendix~\ref{app:lp}.
Figure~\ref{fig:scaling} shows that average balanced accuracy under LP improves with model size from 0.584 at 18M to 0.684 at 396M. 
The largest gain occurs between 48M and 194M (+0.060), which coincides with the largest jump in hidden dimension (512 $\rightarrow$ 896). 
The smaller improvement from 194M to 396M (+0.023) suggests diminishing returns at the current corpus size. 
A similar pattern is observed under LPFT in Table~\ref{tab:main}, where the average increases from 0.669 to 0.767 across the four scales, with a smaller gain from Base to Large. 
In addition, SPERA-Small (48M) achieves performance comparable to that of REVE-Base (69M), and SPERA-Base (194M) exceeds REVE-Large (408M) by +0.013 under LP despite using less than half the parameters.

\begin{figure}[t]
    \centering
    \includegraphics[width=.8\textwidth]{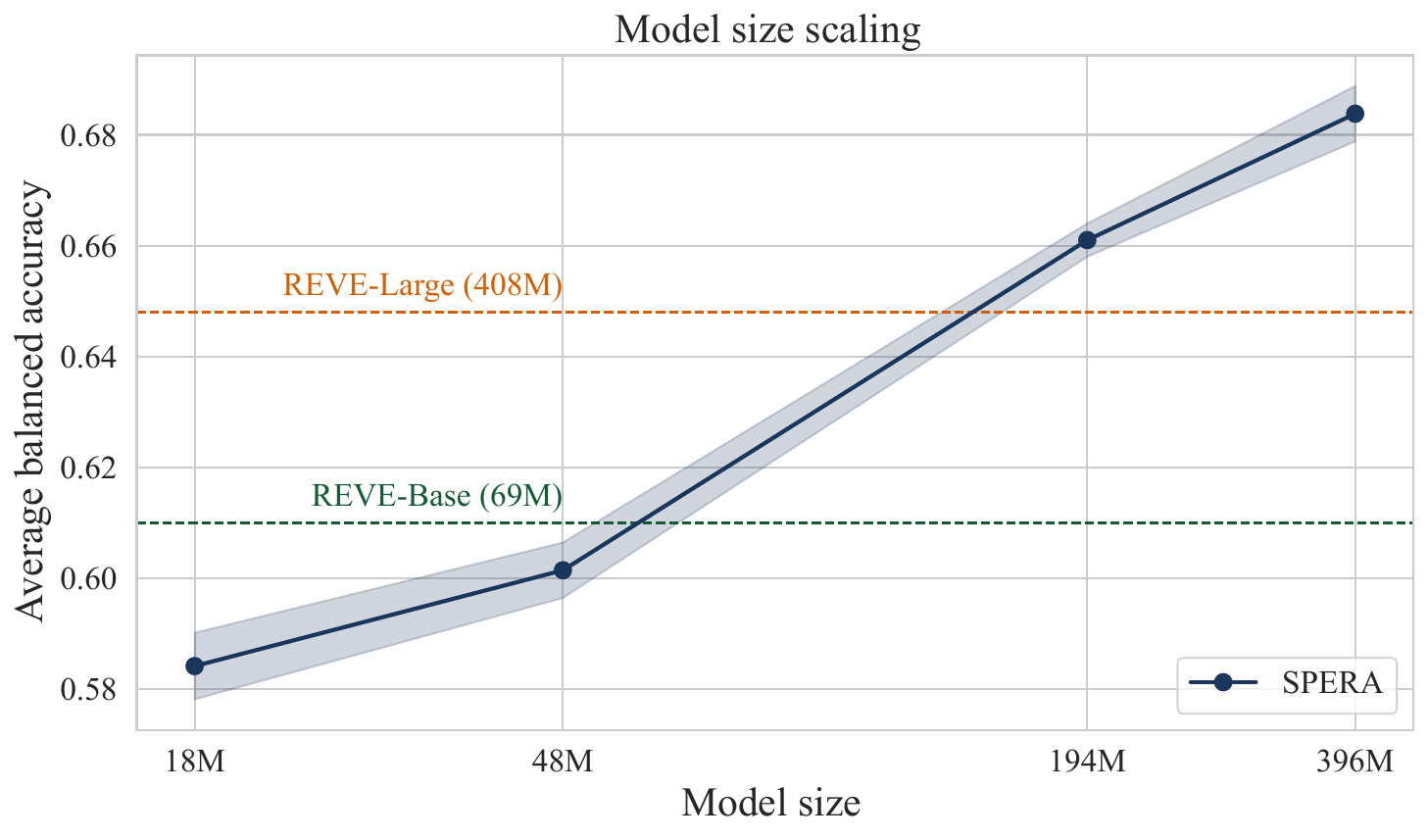}
    \caption{Model scaling under LP. Mean balanced accuracy across nine downstream tasks is shown for SPERA models ranging from 18M to 396M parameters. The shaded region denotes one standard deviation across five random seeds. Horizontal lines indicate the corresponding LP results of REVE-Base (69M) and REVE-Large (408M).}
    \label{fig:scaling}
\end{figure}

\subsection{Ablation study and efficiency analysis}
\label{sec:ablation}

\begin{table}[t]
    \centering
    \caption{Ablation studies on the encoder architecture, RSReg, and electrode geometry encoding. All models are based on SPERA-Small (48M) and evaluated using LP on motor imagery, sleep staging, and EEG abnormality detection tasks. Results are reported as mean balanced accuracy $\pm$ standard deviation over five random seeds. The best result per column is in \textbf{bold}; the second-best result is \underline{underlined}.}
    % \vspace{2mm}
    \label{tab:ablation}
    \resizebox{\textwidth}{!}{
    \begin{tabular}{lcccc}
        \toprule
        \textbf{Variant} & \textbf{PhysioNet-MI} & \textbf{ISRUC} & \textbf{TUAB} & \textbf{Average}\\
        \midrule
        \multicolumn{5}{l}{\textbf{Encoder architecture}}\\
        Full attention only & $0.4450 \pm 0.0124$ & $0.7097 \pm 0.0218$ & $0.7749 \pm 0.0024$ & $0.6432 \pm 0.0084$ \\
        Factorized attention only & $0.4833 \pm 0.0091$ & $0.7153 \pm 0.0187$ & $0.7865 \pm 0.0066$ & $0.6617 \pm 0.0073$ \\
        Hybrid + late full & $0.4871 \pm 0.0181$ & $0.6973 \pm 0.0155$ & $0.7860 \pm 0.0045$ & $0.6568 \pm 0.0081$ \\
        \midrule
        \multicolumn{5}{l}{\textbf{RSReg}}\\
        No RSReg & 0.5182 $\pm$ 0.0184 & $0.6994 \pm 0.0205$ & $0.7759 \pm 0.0046$ & $0.6645 \pm 0.0093$ \\
        \midrule
        \multicolumn{5}{l}{\textbf{Electrode geometry encoding}}\\
        Coordinate embedding only & $0.4821 \pm 0.0150$ & $0.7009 \pm 0.0268$ & $0.7744 \pm 0.0056$ & $0.6525 \pm 0.0104$ \\
        Legendre bias only & \underline{0.5193 $\pm$ 0.0106} & \textbf{0.7320 $\pm$ 0.0182} & \textbf{0.7929 $\pm$ 0.0034} & \textbf{0.6814 $\pm$ 0.0071} \\
        Legendre anchor encoding only & $0.5071 \pm 0.0197$ & $0.7137 \pm 0.0201$ & $0.7855 \pm 0.0028$ & $0.6688 \pm 0.0094$ \\
        Learnable pairwise MLP & $0.4463 \pm 0.0203$ & $0.6679 \pm 0.0264$ & $0.7671 \pm 0.0067$ & $0.6271 \pm 0.0113$ \\
        REVE spatial encoding & $0.4921 \pm 0.0174$ & $0.6910 \pm 0.0212$ & $0.7607 \pm 0.0091$ & $0.6479 \pm 0.0091$ \\
        Gaussian (geodesic) & $0.4512 \pm 0.0161$ & $0.6977 \pm 0.0225$ & $0.7705 \pm 0.0089$ & $0.6398 \pm 0.0101$ \\
        \midrule
        SPERA-Small & \textbf{0.5227 $\pm$ 0.0112} & \underline{0.7206 $\pm$ 0.0240} & \underline{0.7869 $\pm$ 0.0062} & \underline{0.6767 $\pm$ 0.0110} \\
        \bottomrule
    \end{tabular}
    }
\end{table}

\begin{table}[t]
    \centering
    \caption{Computational efficiency and memory profiling using the SPERA-Small model. All measurements are conducted on a single NVIDIA RTX 5000 Ada GPU with a batch size of 1 and a sequence length $N=$ 3,840 (64 channels $\times$ 60 patches). `Mem' denotes peak GPU memory.}
    \label{tab:efficiency_profiling}
    % \vspace{2mm}
    \resizebox{\columnwidth}{!}{
    \begin{tabular}{llcccc}
        \toprule
        \textbf{Architecture} & \textbf{Legendre prior} & \textbf{Params (M)} & \textbf{Matmul FLOPs (G)} & \textbf{Mem (GB)} & \textbf{Throughput (tok/s)} \\
        \midrule
        \multirow{2}{*}{Full} & Bias & 37.9 & 653.0 & 2.51 & 20,260 \\
        & Anchor encoding & 38.7 & 653.1 & 1.61 & 75,769 \\
        \midrule
        \multirow{2}{*}{Factorized}  & Bias & 50.6 & 399.0 & 1.70 & 97,043 \\
        & Anchor encoding & 51.3 & 399.1 & 1.78 & 90,714 \\
        \midrule
        \multirow{3}{*}{Hybrid} & Bias & 47.4 & 462.5 & 2.26 & 52,263 \\
        & Anchor encoding & 48.2 & 462.6 & 1.66 & 96,165 \\
        & Off & 47.4 & 462.5 & 1.59 & 100,259 \\
        \midrule
        SPERA-Small & \makecell[l]{Full (anchor encoding) \\+ factorized (bias)} & 47.6 & 462.5 & 1.95 & 92,866 \\
        \bottomrule
    \end{tabular}
    }
\end{table} 

To understand the contribution of each component, we conduct ablation studies on the encoder architecture, RSReg, and electrode geometry encoding. 
Given the substantial computational cost of repeatedly pretraining the SPERA-Base model (194M), we adopt the SPERA-Small model (48M) as a proxy, evaluated on motor imagery, sleep staging, and EEG abnormality detection tasks using LP. 
We jointly analyze accuracy (Table~\ref{tab:ablation}) and computational efficiency (Table~\ref{tab:efficiency_profiling}) to assess both representational quality and viability under realistic compute budgets.

\paragraph{Component ablations}
We ablate three components: the hybrid attention backbone, RSReg, and the Legendre-based spatial priors.
Neither full nor factorized attention alone matches our periodic hybrid design, and a variant that concentrates full-attention blocks at the end of the network (``Hybrid + late full'' in Table~\ref{tab:ablation}) improves over full-only but still underperforms periodic interleaving. 
This suggests that periodic cross-axis interaction benefits EEG modeling. 
Removing RSReg lowers average accuracy by 0.012, with the largest drop on ISRUC ($-$0.021), suggesting a benefit of frequency-aware supervision.
The Legendre prior improves average accuracy by 0.024 over coordinate embeddings alone (0.6767 vs. 0.6525), and outperforms both REVE's spatial encoding substituted for our Legendre prior (0.6479) and a Gaussian bias on geodesic electrode distance (0.6398). 
The lower performance of the pairwise MLP suggests that the structured Legendre parameterization is beneficial under the present training setup.
Further geometry variants are reported in Appendix~\ref{app:prior:ablation}.
We next examine the computational trade-offs that determine which Legendre implementation is used in each attention type.

\paragraph{Accuracy-efficiency trade-off}
Factorized attention reduces the matrix-multiplication cost substantially at our sequence length ($N$= 3,840), from 653 GFLOPs for full attention to 399 GFLOPs, despite having more parameters due to the additional axis-specific projections (50.6M vs.\ 37.9M). 
The hybrid architecture lies between the two at 462.5 GFLOPs, reflecting its combination of factorized blocks with periodic full-attention blocks. 
Here, matrix-multiplication FLOPs include linear projections and the attention matrix multiplications, with one multiply-accumulate counted as two FLOPs.

The Legendre implementation has little effect on the matrix-multiplication cost but strongly affects realized throughput. 
Under full attention, the pairwise Legendre bias requires an $N \times N$ score-bias matrix and prevents the fused attention path used in our implementation (i.e., FlashAttention via PyTorch SDPA), reducing throughput to 20,260 tok/s despite nearly identical matrix-multiplication FLOP counts for the bias- and anchor-based implementations (653.0 vs.\ 653.1 GFLOPs). 
The anchor encoding instead injects geometry features into the queries and keys while preserving the fused
attention path, reaching 75,769 tok/s.
In factorized attention, the smaller attention matrices make the explicit pairwise bias practical, achieving 97,043 tok/s compared with 90,714 tok/s for the anchor encoding.

These observations motivate our final design: we use the anchor encoding in full-attention blocks to preserve throughput and the pairwise Legendre bias in factorized layers, where its computational cost is small. 
The resulting model achieves accuracy comparable to that of the bias-only hybrid configuration while providing 1.78$\times$ higher throughput (92,866 vs.\ 52,263 tok/s) and 14\% lower peak memory (1.95 vs.\ 2.26 GB), at the cost of a 0.005 decrease in average balanced accuracy.

\subsection{Additional analyses}
We provide further analyses in the appendix, including controlled comparisons of the pretraining objective and corpus (Appendix~\ref{app:ablation:objective}), temporal positional encoding and masking strategy (Appendices~\ref{app:ablation:pe} and \ref{app:masking}), analyses of the Legendre spherical prior (Appendix~\ref{app:prior}), few-shot performance (Appendix~\ref{app:fewshot}), noise and ocular-artifact robustness (Appendix~\ref{app:noise}), and learned representation analysis (Appendix~\ref{app:representation_analysis}).

\section{Limitations and future work}
\label{sec:limitations}
While our scaling analysis covers four model sizes from 18M to 396M parameters, we did not investigate scaling at the billion-parameter scale due to limited computational resources. 
Furthermore, the diminishing returns observed beyond 194M suggest that further gains may require expanded data rather than parameters alone, but we did not systematically disentangle the contributions of corpus size, model capacity, and architectural priors to the observed scaling behavior. 
A systematic characterization of these factors remains an important direction for future investigation.
Beyond scale, the spherical prior assumes surface electrodes; extending it to intracranial recordings would require incorporating radial information, as discussed in Appendix~\ref{app:prior:ieeg}.

We evaluate SPERA on nine diverse downstream tasks spanning clinical, cognitive, and BCI domains; however, the tasks are focused on classification problems over fixed-duration EEG segments ranging from 3 to 30 seconds.
Further validation is therefore needed to assess generalization to continuous decoding, regression targets, and long-horizon clinical prediction. 
In addition, demographic metadata such as age, sex, ethnicity, and clinical status are not consistently available across our pretraining datasets, limiting a comprehensive fairness analysis of the learned representations across subpopulations. 
Addressing these gaps will be important before clinical or consumer deployment.

\section{Conclusion}
\label{conclusion}
We present SPERA, a foundation model for EEG that learns representations through latent-space prediction. 
SPERA encodes scalp electrode geometry into attention through a Legendre-polynomial spatial prior, supported by a hybrid attention backbone and RSReg.

Across nine downstream tasks, SPERA-Base exceeds the strongest baseline (REVE-Base; Table~\ref{tab:main}) by 0.035 in average balanced accuracy. 
The largest significant gains are observed on fine-grained multi-class tasks, including imagined speech and clinical event classification. 
The scaling analysis further shows that SPERA-Base exceeds REVE-Large under LP despite using less than half the parameters. 
Controlled ablations show that latent prediction outperforms a matched reconstruction objective and that the Legendre prior outperforms the alternative geometry encodings we evaluated.

Overall, the results support latent-space prediction, geometry-aware spatial modeling, and frequency-aware supervision as complementary components for EEG representation learning across heterogeneous datasets and electrode montages. 
These findings provide a basis for future work on scalable and transferable EEG representation learning.

\begin{ack}
This work was supported by the Samsung Science \& Technology Foundation Program (SRFC-IT2401-05). The authors declare no competing interests.
\end{ack}

\renewcommand{\bibfont}{\small}
\bibliographystyle{plain}
\bibliography{reference}

%%%%%%%%%%%%%%%%%%%%%%%%%%%%%%%%%%%%%%%%%%%%%%%%%%%%%%%%%%%%
\newpage
\appendix

\section{Computing resources}
\label{app:computing_resources}
To ensure reproducibility, we report the detailed computing environment, as follows:
\begin{itemize}
    \item \textbf{Hardware:} HP Z8 Fury G5 Workstation equipped with an Intel Xeon w5-3425 CPU, 128 GB of RAM, and dual NVIDIA RTX 5000 Ada Generation GPUs.
    \item \textbf{Software \& environment:} The environment is set up on Ubuntu 24.04.4 LTS and Python 3.10.20, supported by CUDA Toolkit 13.0 and cuDNN 9.15 for hardware acceleration. Model implementation and training are built upon PyTorch 2.10.0 and the Hugging Face ecosystem, utilizing \texttt{transformers} 5.3.0 and \texttt{accelerate} 1.13.0. Data preprocessing is performed using MNE-Python 1.11.0.
    \item \textbf{Training efficiency \& cost:} To maximize training efficiency, we employed distributed data parallel (DDP) across both GPUs and utilized Bfloat16 (BF16) mixed-precision training. The total computational cost for pretraining our largest model is approximately 287 GPU hours. 
    \item \textbf{Code \& model availability:} Our codebase is provided through the repository linked in the paper, including training, evaluation, and preprocessing scripts.
    %We have made our codebase and pretrained models publicly available to facilitate reproducibility and further research in the community.
\end{itemize}

\section{Pretraining corpus}
\label{app:ptds}
Datasets are collected from multiple public repositories, including OpenNeuro~\cite{markiewicz2021openneuro}, MOABB~\cite{Aristimunha_Mother_of_all}, PhysioNet~\cite{goldberger2000physiobank}, NSRR~\cite{zhang2018national}, TUH~\cite{obeid2016temple}, Brainclinics (\url{https://brainclinics.com/resources}; see also~\cite{TDBRAIN}), and the Max Planck Institute~\cite{babayan2019mind}. Table~\ref{tab:appendix_b1_summary} reports subject counts adjusted for known cross-dataset overlaps, and Table~\ref{tab:appendix_b2_datasets} lists per-dataset counts. To avoid double-counting, known subject-level overlaps are removed (e.g., overlaps among the Shin2017 series~\cite{shin2017eegnirs}, Lee2019 series~\cite{lee2019openbmi}, and OpenNeuro datasets ds007169~\cite{ds007169}/ds007262~\cite{ds007262}). In addition, to prevent any evaluation leakage, subjects reserved for downstream tasks (TUAB and TUEV) are excluded from the TUH corpus. For HSP~\cite{sun2023human}, we use the small training subset released for the PhysioNet Challenge 2026~\cite{2026Challenge}. Unless otherwise noted, reported hours reflect the pretraining corpus after these exclusions. 

\begin{table}[htbp]
    \centering
    \caption{Summary statistics of the pretraining corpus (aggregated from dataset-level metadata).}
    % \vspace{2mm}
    \label{tab:appendix_b1_summary}
    \begin{tabular}{lrrr}
        \toprule
        \textbf{Grouping} & \textbf{\# Subjects} & \textbf{Hours (h)} & \textbf{\# Datasets} \\
        \midrule
        \multicolumn{4}{l}{\textbf{By category}}\\
        Clinical & 21,916 & 65,387 & 11 \\
        Cognitive & 6,248 & 12,915 & 55 \\ 
        BCI & 884 & 1,362 & 40 \\ % 1143 - lee108, shin29, 7120 = 986
        \midrule
        \multicolumn{4}{l}{\textbf{By channel count}}\\
        3--32 & 22,728 & 66,051 & 40 \\
        33--64 & 1,963 & 2,114 & 42 \\
        $> 64$ & 4,357 & 11,499 & 24 \\
        \midrule
        \multicolumn{4}{l}{\textbf{By segment length}}\\
        10 s & 3,414 & 1,967 & 46 \\
        30 s & 4,236 & 11,378 & 19 \\
        60 s & 21,398 & 66,319 & 41 \\
        \midrule
        Total & 29,048 & 79,664 & 106 \\
        \bottomrule
    \end{tabular}
\end{table}

{\small
\captionsetup{font=normalsize}
\begin{longtable}{p{3.0cm}p{1.6cm}rrr p{2.5cm}}
\caption{Datasets used in the pretraining corpus.}
\label{tab:appendix_b2_datasets}\\
\toprule
\textbf{Dataset} & \textbf{Repository} & \textbf{\# Subjects} & \textbf{Hours (h)} & \textbf{Channels} & \textbf{License} \\
\midrule
\endfirsthead
\toprule
\textbf{Dataset} & \textbf{Repository} & \textbf{\# Subjects} & \textbf{Hours (h)} & \textbf{Channels} & \textbf{License} \\
\midrule
\endhead
\midrule \multicolumn{6}{r}{\textit{Continued on next page}} \\
\endfoot
\bottomrule
\endlastfoot

\multicolumn{6}{l}{\textbf{Clinical}}\\
\addlinespace
TUH~\cite{obeid2016temple} & TUH & 12,299 & 25,785 & Mostly 30--41 & \makecell[lt]{NEDC data use\\agreement}\\TDBRAIN~\cite{TDBRAIN} & Brainclinics & 1,274 & 81.67 & 26 & \makecell[lt]{Brainclinics data\\use agreement}\\
MESA~\cite{MESA_Sleep} & NSRR & 2,056 & 10,568 & 3 & \makecell[lt]{NSRR data use\\agreement}\\
Siena~\cite{PhysioNet-siena-scalp-eeg-1.0.0,detti2020eeg} & PhysioNet & 14 & 141 & 29 & CC BY 4.0\\
I-CARE~\cite{PhysioNet-i-care-2.1, amorim2023international} & PhysioNet & 607 & 22,278 & 19 & CC BY-NC-SA 4.0\\
HSP (subset)~\cite{sun2023human,2026Challenge} & Kaggle & 1,103 & 5,799 & 6 & BDSP license 1.5.0\\
HMS~\cite{HMS_HBAC} & Kaggle & 1,950 & 237 & 8 & CC BY-NC 4.0\\
NMT~\cite{khan2022nmt} & Zenodo & 2,417 & 474 & 19 & CC BY 4.0\\
ds002778~\cite{ds002778} & OpenNeuro & 31 & 2.33 & 32 & CC0\\
ds004902~\cite{ds004902} & OpenNeuro & 71 & 17.62 & 61 & CC0\\
ds007020~\cite{ds007020} & OpenNeuro & 94 & 3.60 & 63 & CC0\\
\addlinespace

\multicolumn{6}{l}{\textbf{Cognitive}}\\
\addlinespace
MPI-Lemon~\cite{babayan2019mind} & \makecell[lt]{Max Planck\\Institute} & 215 & 48.35 & 59 & PDDL\\
THINGS2~\cite{gifford2022large} & FigShare& 10 & 79.25 & 64 & CC BY 4.0\\
ds002680~\cite{ds002680} & OpenNeuro & 14 & 17.85 & 19 & CC0\\
ds002721~\cite{ds002721} & OpenNeuro & 31 & 25.17 & 19 & CC0\\
ds003004~\cite{ds003004} & OpenNeuro & 34 & 47.25 & 134 & CC0\\
ds003690~\cite{ds003690} & OpenNeuro & 75 & 45.13 & 59 & CC0\\
ds003825~\cite{ds003825} & OpenNeuro & 50 & 45.75 & 63 & CC0\\
ds003885~\cite{ds003885} & OpenNeuro & 24 & 26.80 & 128 & CC0\\
ds003969~\cite{ds003969} & OpenNeuro & 98 & 65.65 & 64 & CC0\\
ds004043~\cite{ds004043} & OpenNeuro & 20 & 18.03 & 63 & CC0\\
ds004147~\cite{ds004147} & OpenNeuro & 12 & 9.50 & 31 & CC0\\
ds004148~\cite{ds004148} & OpenNeuro & 60 & 60.00 & 64 & CC0\\
ds004264~\cite{ds004264} & OpenNeuro & 21 & 7.27 & 31 & CC0\\
ds004315~\cite{ds004315} & OpenNeuro & 50 & 21.04 & 60 & CC0\\
ds004356~\cite{ds004356} & OpenNeuro & 22 & 46.03 & 34 & CC0\\
ds004395~\cite{ds004395} & OpenNeuro & 364 & 8,687 & 128 & CC0\\
ds004408~\cite{ds004408} & OpenNeuro & 19 & 18.68 & 128 & CC0\\
ds004477~\cite{ds004477} & OpenNeuro & 9 & 13.48 & 79 & CC0\\
ds004561~\cite{ds004561} & OpenNeuro & 23 & 11.35 & 61 & CC0\\
ds004572~\cite{ds004572} & OpenNeuro & 52 & 47.37 & 58 & CC0\\
ds004582~\cite{ds004582} & OpenNeuro & 73 & 33.65 & 59 & CC0\\
ds004603~\cite{ds004603} & OpenNeuro & 37 & 30.61 & 64 & CC0\\
ds004706~\cite{ds004706} & OpenNeuro & 34 & 468 & 128 & CC0\\
ds004816~\cite{ds004816} & OpenNeuro & 20 & 10.50 & 63 & CC0\\
ds004951~\cite{ds004951} & OpenNeuro & 11 & 25.75 & 63 & CC0\\
ds005095~\cite{ds005095} & OpenNeuro & 48 & 16.48 & 63 & CC0\\
ds005121~\cite{ds005121} & OpenNeuro & 34 & 40.17 & 64 & CC0\\
ds005189~\cite{ds005189} & OpenNeuro & 30 & 19.28 & 61 & CC0\\
ds005273~\cite{ds005273} & OpenNeuro & 33 & 51.17 & 63 & CC0\\
ds005410~\cite{ds005410} & OpenNeuro & 81 & 22.94 & 63 & CC0\\
ds005505~\cite{shirazi2024hbneeg} & OpenNeuro & 136 & 105 & 128 & CC BY-SA 4.0\\
ds005506~\cite{shirazi2024hbneeg} & OpenNeuro & 150 & 115 & 128 & CC BY-SA 4.0\\
ds005507~\cite{shirazi2024hbneeg} & OpenNeuro & 184 & 142 & 128 & CC BY-SA 4.0\\
ds005508~\cite{shirazi2024hbneeg} & OpenNeuro & 324 & 137 & 128 & CC BY-SA 4.0\\
ds005509~\cite{shirazi2024hbneeg} & OpenNeuro & 330 & 251 & 128 & CC BY-SA 4.0\\
ds005510~\cite{shirazi2024hbneeg} & OpenNeuro & 134 & 92.20 & 128 & CC BY-SA 4.0\\
ds005511~\cite{shirazi2024hbneeg} & OpenNeuro & 381 & 250 & 128 & CC BY-SA 4.0\\
ds005512~\cite{shirazi2024hbneeg} & OpenNeuro & 257 & 164 & 128 & CC BY-SA 4.0\\
ds005514~\cite{shirazi2024hbneeg} & OpenNeuro & 295 & 183 & 128 & CC BY-SA 4.0\\
ds005515~\cite{shirazi2024hbneeg} & OpenNeuro & 533 & 148 & 128 & CC BY-SA 4.0\\
ds005516~\cite{shirazi2024hbneeg} & OpenNeuro & 430 & 245 & 128 & CC BY-SA 4.0\\
ds005594~\cite{ds005594} & OpenNeuro & 16 & 12.17 & 64 & CC0\\
ds005620~\cite{ds005620} & OpenNeuro & 21 & 16.25 & 62 & CC0\\
ds005662~\cite{ds005662} & OpenNeuro & 80 & 79.85 & 64 & CC0\\
ds006018~\cite{ds006018} & OpenNeuro & 127 & 50.39 & 26 & CC0\\
ds006434~\cite{ds006434} & OpenNeuro & 66 & 120 & 32 & CC0\\
ds006480~\cite{ds006480} & OpenNeuro & 68 & 69.98 & 65 & CC0\\
ds006850~\cite{ds006850} & OpenNeuro & 63 & 78.30 & 64 & CC0\\
ds006861~\cite{ds006861} & OpenNeuro & 120 & 100 & 32 & CC0\\
ds006866~\cite{ds006866} & OpenNeuro & 148 & 122 & 62 & CC0\\
ds006921~\cite{ds006921} & OpenNeuro & 38 & 16.01 & 63 & CC0\\
ds006923~\cite{ds006923} & OpenNeuro & 140 & 33.85 & 128 & CC0\\
ds007172~\cite{ds007172} & OpenNeuro & 100 & 47.89 & 32 & CC0\\
ds007176~\cite{ds007176} & OpenNeuro & 45 & 25.08 & 58 & CC0\\
HBN\_NC~\cite{shirazi2024hbneeg} & \makecell[lt]{Child Mind\\Institute} & 458 & 251 & 128 & CC BY-NC-SA 4.0\\

\addlinespace
\multicolumn{6}{l}{\textbf{BCI}}\\
\addlinespace
ds003846~\cite{ds003846} & OpenNeuro & 19 & 22.38 & 63 & CC0\\
ds004152~\cite{ds004152} & OpenNeuro & 21 & 11.27 & 31 & CC0\\
ds004284~\cite{ds004284} & OpenNeuro & 18 & 9.43 & 128 & CC0\\
ds004324~\cite{ds004324} & OpenNeuro & 26 & 19.18 & 28 & CC0\\
ds004357~\cite{ds004357} & OpenNeuro & 16 & 11.28 & 128 & CC0\\
ds004817~\cite{ds004817} & OpenNeuro & 20 & 11.88 & 63 & CC0\\
ds005089~\cite{ds005089} & OpenNeuro & 36 & 62.76 & 63 & CC0\\
ds005262~\cite{ds005262} & OpenNeuro & 12 & 24.62 & 8 & CC0\\
ds005586~\cite{ds005586} & OpenNeuro & 23 & 33.49 & 60 & CC0\\
ds005697~\cite{ds005697} & OpenNeuro & 51 & 70.67 & 64 & CC0\\
ds007169~\cite{ds007169} & OpenNeuro & 18 & 4.90 & 19 & CC0\\
ds007262~\cite{ds007262} & OpenNeuro & 18 & 4.43 & 19 & CC0\\
AlexMI~\cite{AlexMI} & MOABB & 8 & 1.09 & 16 & CC BY-SA 4.0\\
Beetl2021\_A~\cite{Beetl2021_A} & MOABB & 3 & 1.00 & 63 & CC BY 4.0 \\
BI2014a~\cite{BI2014a} & MOABB & 64 & 12.32 & 16 & CC BY 4.0 \\
BI2014b~\cite{BI2014b} & MOABB & 38 & 2.31 & 32 & CC BY 4.0 \\
BI2015a~\cite{BI2015a} & MOABB & 43 & 11.48 & 32 & CC BY 4.0 \\
BI2015b~\cite{BI2015b} & MOABB & 44 & 25.87 & 32 & CC BY 4.0 \\
BNCI2003\_004~\cite{BNCI2003_004} & MOABB & 5 & 3.97 & 118 & CC BY 4.0 \\
BNCI2014\_004~\cite{leeb2007brain} & MOABB & 9 & 26.19 & 3 & CC BY-ND 4.0 \\
BNCI2014\_008~\cite{BNCI2014_008} & MOABB & 8 & 3.11 & 8 & CC BY-NC-ND 4.0 \\
BNCI2014\_009~\cite{BNCI2014_009} & MOABB & 10 & 1.58 & 16 & CC BY-NC-ND 4.0 \\
BNCI2015\_001~\cite{BNCI2015_001} & MOABB & 12 & 16.66 & 13 & CC BY-NC-ND 4.0 \\
BNCI2015\_003~\cite{BNCI2015_003} & MOABB & 10 & 0.91 & 8 & CC BY-NC-ND 4.0 \\
BNCI2015\_004~\cite{BNCI2015_004} & MOABB & 9 & 13.72 & 30 & CC BY-NC-ND 4.0 \\
Cho2017~\cite{Cho2017} & MOABB & 52 & 20.46 & 64 & CC BY 4.0 \\
Dreyer2023~\cite{Dreyer2023} & MOABB & 87 & 68.12 & 27 & CC BY 4.0 \\
EPFLP300~\cite{EPFLP300} & MOABB & 8 & 2.79 & 32 & Not specified \\
Kalunga2016~\cite{Kalunga2016} & MOABB & 12 & 2.56 & 8 & CC BY 4.0 \\
Lee2019ERP~\cite{lee2019openbmi} & MOABB & 54 & 57.85 & 62 & GPL-3.0\\
Lee2019MI~\cite{lee2019openbmi} & MOABB & 54 & 42.81 & 62 & GPL-3.0\\
Lee2019SSVEP~\cite{lee2019openbmi} & MOABB & 54 & 40.45 & 62 & GPL-3.0\\
Ofner2017~\cite{Ofner2017} & MOABB & 15 & 13.33 & 61 & CC BY 4.0\\
Schirrmeister2017~\cite{Schirrmeister2017} & MOABB & 14 & 26.44 & 128 & CC BY 4.0 \\
Shin2017A~\cite{shin2017eegnirs} & MOABB & 29 & 14.41 & 30 & GPL-3.0\\
Shin2017B~\cite{shin2017eegnirs} & MOABB & 29 & 14.43 & 30 & GPL-3.0\\
Sosulski2019~\cite{Sosulski2019} & MOABB & 13 & 18.30 & 31 & CC BY-SA 4.0\\
Stieger2021~\cite{Stieger2021} & MOABB & 62 & 615 & 60 & CC BY-NC 4.0\\
Weibo2014~\cite{Weibo2014} & MOABB & 10 & 12.44 & 60 & CC0\\
Zhou2016~\cite{Zhou2016} & MOABB & 4 & 6.24 & 14 & CC BY 4.0\\
\end{longtable}
}

\section{Pretraining details}
\subsection{Detailed hyperparameters}
\label{app:impl}
In this section, we present the detailed hyperparameters for pretraining our models. Table~\ref{tab:shared_param} summarizes the universal settings shared across all model sizes, including the core architecture, optimization schedule, and augmentation parameters. Table~\ref{tab:model_param} details the specific architectural dimensions and learning rates scaled for the 18M, 48M, 194M, and 396M models.

\begin{table}[h]
    \centering
    \caption{Hyperparameters for the SPERA model pretraining.}
    % \vspace{2mm}
    \label{tab:shared_param}
    \begin{tabular}{lc}
        \toprule
        \textbf{Hyperparameter} & \textbf{Value} \\
        \midrule
        \multicolumn{2}{l}{\textbf{EEG input \& tokenization}} \\
        Sampling rate       &   200 Hz \\
        Patch size          &   1.0 sec \\
        \midrule
        \multicolumn{2}{l}{\textbf{Masking \& student augmentation}} \\
        Temporal/spatial mask probability   &   0.80 / 0.80   \\
        Temporal mask ratio (min/max)       &   0.15 / 0.35 \\
        Spatial mask ratio (min/max)    &   0.10 / 0.30   \\
        Amplitude gain (min/max)        &   0.80 / 1.20   \\
        Gaussian noise std (min/max)    &   0.00 / 0.03  \\
        Coordinate jitter (prob/std)    &   0.50 / 0.05  \\
        \midrule
        \multicolumn{2}{l}{\textbf{Architecture}} \\
        MLP type / ratio    &   SwiGLU / 4.0 \\
        Predictor layers    &   2 \\
        Encoder architecture &  Hybrid (full attn every 4th layer) \\
        Normalization       &   Layer normalization \\
        Dropout (residual/attention) & 0.0 \\
        Temporal embedding  &   RoPE ($\theta=$ 10000)  \\
        Legendre degree ($L$)     &   8 \\
        Legendre anchor points ($M$) &   32 \\
        \midrule
        \multicolumn{2}{l}{\textbf{Optimization}} \\
        Optimizer           &   AdamW ($\beta_1=$ 0.9, $\beta_2=$ 0.95) \\
        Weight decay        &   0.05    \\
        Total steps         &   47{,}000  \\
        Input tokens per update (nominal)   &   393{,}216 \\
        \quad - Context / target tokens &   262{,}144 / 131{,}072 \\
        LR scheduler        &   Warmup-cosine annealing \\
        Warmup ratio        &   0.05 \\
        EMA schedule        &   Cosine annealing (0.996 $\to$ 0.9999) \\
        \midrule
        \multicolumn{2}{l}{\textbf{RSReg}} \\
        Weight schedule     &   Delayed linear ramp-decay \\
        Weight bounds       &   0 $\to$ 0.02 $\to$ 0.002 \\
        Schedule steps      &   5\% (ramp start), 10\% (peak), 50\% (decay start) \\
        Projection dimension & 128 \\
        Temperature ($\tau_z, \tau_s$) & 0.1, 0.1 \\
        \bottomrule
    \end{tabular}
\end{table}

\begin{table}[h]
    \centering
    \caption{Scale-specific architecture and optimization hyperparameters. Attention and predictor head counts and hidden dimensions are identical within each model size. Learning rates are scaled according to the model size.}
    % \vspace{2mm}
    \label{tab:model_param}
    \begin{tabular}{lcccccc}
        \toprule
        \textbf{Model size} & \textbf{Params} & \textbf{Layers} & \textbf{Hidden dim} & \textbf{Attention heads} & \textbf{Peak LR} \\
        \midrule
        Tiny        &   18M     &   8       &   384         &   4               &   $4.0 \times 10^{-4}$    \\
        Small       &   48M     &   12      &   512         &   8               &   $3.5 \times 10^{-4}$    \\
        Base        &   194M    &   16      &   896         &   14              &   $3.0 \times 10^{-4}$    \\
        Large       &   396M    &   25      &   1024        &   16              &   $2.5 \times 10^{-4}$    \\
        \bottomrule
    \end{tabular}
\end{table}

\subsection{Pretraining token accounting and loss curve}
\label{app:token}

We train SPERA for 47{,}000 optimization steps, corresponding to approximately two epochs over the pretraining corpus.
Optimization steps are triggered adaptively: microbatches accumulate until the cumulative target-token count reaches the per-update budget of 131{,}072 tokens (Table~\ref{tab:shared_param}), so the number of microbatches per update varies. 

\begin{figure}[tbp]
    \centering
    \includegraphics[width=0.8\textwidth]{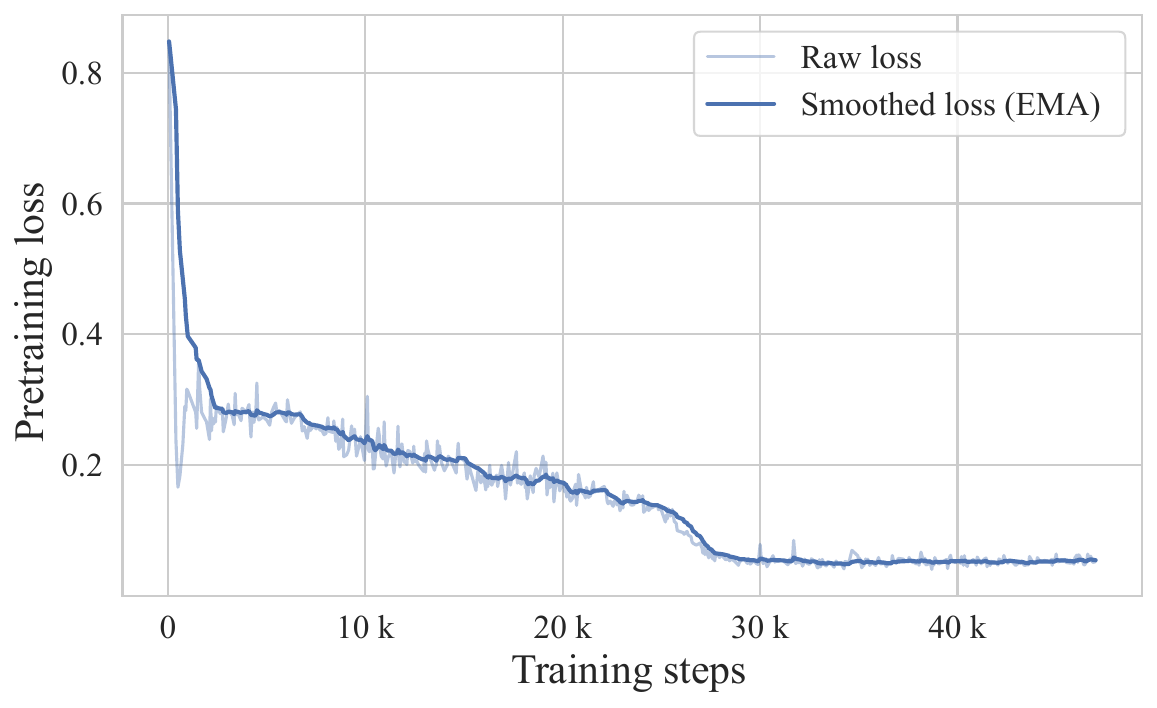}
    \caption{Pretraining loss curve of the SPERA-Base (194M) model. The model is trained for approximately 47,000 steps ($\sim$2 epochs of the pretraining corpus). The raw per-step loss (light) is shown together with an exponential moving average for readability (solid); this plot-level smoothing is distinct from the EMA used to update the teacher encoder.}
    \label{fig:pretraining_loss}
\end{figure}

Figure~\ref{fig:pretraining_loss} shows the pretraining loss curve of the SPERA-Base (194M) model. 
Each epoch realizes $\sim$9.45B input tokens ($\sim$6.25B context, $\sim$3.20B target), or $\sim$98.4\% of the $\sim$9.60B nominal token count derived from corpus statistics (Table~\ref{tab:appendix_b1_summary}).
The remaining $\sim$1.6\% corresponds to rare $(C, P_{t})$ shape buckets that do not accumulate a full microbatch within the dataloader's pending window and are dropped by the shape-aware batcher. 
The residual is small because the unit that is dropped is a rare shape bucket rather than a dataset: even our smallest dataset ($\sim$0.91 h, 8 channels) produces $\sim$26K tokens, exceeding the per-microbatch budget. 
The resulting loss curve (Figure~\ref{fig:pretraining_loss}) converges smoothly and stabilizes around 0.05 by the end of training.

\section{Detailed downstream task results}
\label{app:down}
In this section, we describe each downstream dataset, its preprocessing procedure, and the full experimental results.
For all datasets, we follow the preprocessing pipelines of prior works~\cite{yang2023biot,jiang2024large,wang2024cbramod,ouahidi2025reve} to ensure direct comparison with baseline models. 
While we report balanced accuracy as the primary metric in the main text, we additionally report task-specific metrics commonly used in prior EEG foundation model evaluations~\cite{yang2023biot,jiang2024large,wang2024cbramod,ouahidi2025reve}. 
For multi-class classification tasks, we report Cohen's $\kappa$ and weighted F1-score in addition to balanced accuracy. 
For binary classification tasks, we report AUC-PR and AUROC in addition to balanced accuracy. 
SPERA results are reported as mean $\pm$ standard deviation over five random seeds; baseline statistics follow the evaluation protocols of the source publications.
Each downstream task table includes only baseline models whose original studies report results on that task.
A dash indicates that the corresponding model does not report that dataset or metric.
Unless otherwise specified, all results are obtained using the SPERA-Base (194M) model.

\subsection{Imagined speech classification}

\begin{table}[t]
    \centering
    \caption{The results of different methods on imagined speech classification (BCIC2020-3, 5-class).}
    \label{tab:bcic2020_3}
    % \vspace{2mm}
    \begin{tabular}{lccc}
        \toprule
        \textbf{Methods} & \textbf{Balanced accuracy} & \textbf{Cohen's $\kappa$} & \textbf{Weighted F1} \\
        \midrule
        EEGNet          & $0.4413 \pm 0.0096$ & $0.3016 \pm 0.0123$ & $0.4413 \pm 0.0102$ \\
        EEGConformer    & $0.4506 \pm 0.0133$ & $0.3133 \pm 0.0183$ & $0.4488 \pm 0.0154$ \\
        SPaRCNet        & $0.4426 \pm 0.0156$ & $0.3033 \pm 0.0233$ & $0.4420 \pm 0.0108$ \\
        ContraWR        & $0.4257 \pm 0.0162$ & $0.3078 \pm 0.0218$ & $0.4407 \pm 0.0182$ \\
        CNN-Transformer & $0.4533 \pm 0.0092$ & $0.3166 \pm 0.0118$ & $0.4506 \pm 0.0127$ \\
        FFCL            & $0.4678 \pm 0.0197$ & $0.3301 \pm 0.0359$ & $0.4689 \pm 0.0205$ \\
        ST-Transformer  & $0.4126 \pm 0.0122$ & $0.2941 \pm 0.0159$ & $0.4247 \pm 0.0138$ \\
        \midrule
        BIOT            & $0.4920 \pm 0.0086$ & $0.3650 \pm 0.0176$ & $0.4917 \pm 0.0079$ \\
        LaBraM-Base     & $0.5060 \pm 0.0155$ & $0.3800 \pm 0.0242$ & $0.5054 \pm 0.0205$ \\
        CBraMod         & $0.5373 \pm 0.0108$ & $0.4216 \pm 0.0163$ & $0.5383 \pm 0.0096$ \\
        REVE-Base       & $0.5635 \pm 0.0123$ & $0.4543 \pm 0.0154$ & $0.5633 \pm 0.0124$ \\
        CSBrain         & $0.6004 \pm 0.0187$ & $0.5006 \pm 0.0233$ & $0.6003 \pm 0.0192$ \\
        CodeBrain       & $0.6101 \pm 0.0052$ & $0.5127 \pm 0.0065$ & $0.6101 \pm 0.0053$ \\
        EEGMamba        & $0.5305 \pm 0.0109$ & $0.4183 \pm 0.0157$ & $0.5347 \pm 0.0082$ \\
        \midrule
        \textbf{SPERA}   & \textbf{0.6504 $\pm$ 0.0122} & \textbf{0.5630 $\pm$ 0.0152} & \textbf{0.6498 $\pm$ 0.0129}  \\
        \bottomrule
    \end{tabular}
\end{table}

BCIC2020-3~\cite{jeong20222020} is a 5-class imagined speech dataset. 
Fifteen subjects are instructed to silently imagine pronouncing one of five phrases (``hello'', ``help me'', ``stop'', ``thank you'', and ``yes'') without articulation. 
EEG signals are recorded from 64 channels at 256 Hz, and resampled to 200 Hz. 
Each trial consists of a 3-second segment, yielding 6{,}000 samples in total with predefined train/validation/test splits of 60/10/10 trials per class. 

As shown in Table~\ref{tab:bcic2020_3}, SPERA achieves 0.6504 balanced accuracy, with Cohen's $\kappa$ of 0.5630 and weighted F1 of 0.6498.

\subsection{Motor imagery classification}
BCIC-IV-2a~\cite{tangermann2012review} is a 4-class motor imagery dataset comprising nine subjects performing left hand, right hand, both feet, and tongue imagery. 
EEG signals are recorded across two sessions using 22 electrodes at 250 Hz. 
Each trial is cropped to the 2--6 s segment, bandpass filtered to 0.5--99.5 Hz, and resampled to 200 Hz, yielding 5{,}184 4-second samples. 
Following prior work~\cite{ouahidi2025reve}, Euclidean alignment is additionally applied. 
We adopt a subject-independent split, with subjects 1--5, 6--7, and 8--9 used for training, validation, and test, respectively.

PhysioNet-MI~\cite{goldberger2000physiobank,schalk2004bci2000} is a 4-class motor imagery dataset recorded from 64 channels at 160 Hz, covering left fist, right fist, both fists, and both feet imagery. 
We extract 4-second segments yielding 9{,}837 samples, and apply a subject-independent split with subjects 1--70, 71--89, and 90--109 for training, validation, and test, respectively. 
All recordings are loaded at 128 Hz with a 64 Hz low-pass filter and subsequently resampled to 200 Hz.

The detailed experimental results for both motor imagery tasks are summarized in Table~\ref{tab:motor_imagery}.
SPERA improves over REVE-Base on both BCIC-IV-2a ($+$0.0420) and PhysioNet-MI ($+$0.0246) in balanced accuracy, with similar trends in the other two metrics. 
The improvement is consistent across the two datasets despite their differing channel counts (22 vs.\ 64) and acquisition protocols.

\begin{table}[htbp]
    \centering
    \caption{The results of different methods on motor imagery classification.}
    \label{tab:motor_imagery}
    % \vspace{2mm}
    \resizebox{\textwidth}{!}{
    \begin{tabular}{lcccccc}
        \toprule
        \multirow{2}{*}{\textbf{Methods}} & \multicolumn{3}{c}{\textbf{BCIC-IV-2a, 4-class}} & \multicolumn{3}{c}{\textbf{PhysioNet-MI, 4-class}} \\
        \cmidrule(lr){2-4} \cmidrule(lr){5-7}
        & \textbf{Balanced accuracy} & \textbf{Cohen's $\kappa$} & \textbf{Weighted F1} & \textbf{Balanced accuracy} & \textbf{Cohen's $\kappa$} & \textbf{Weighted F1} \\
        \midrule
        EEGNet          & $0.4482 \pm 0.0094$ & $0.2693 \pm 0.0120$ & $0.4226 \pm 0.0108$ & $0.5814 \pm 0.0125$ & $0.4468 \pm 0.0199$ & $0.5796 \pm 0.0115$ \\
        EEGConformer    & $0.4696 \pm 0.0106$ & $0.2924 \pm 0.0141$ & $0.4533 \pm 0.0128$ & $0.6049 \pm 0.0104$ & $0.4736 \pm 0.0171$ & $0.6062 \pm 0.0095$ \\
        SPaRCNet        & $0.4635 \pm 0.0117$ & $0.2847 \pm 0.0147$ & $0.4432 \pm 0.0120$ & $0.5932 \pm 0.0152$ & $0.4564 \pm 0.0234$ & $0.5937 \pm 0.0147$ \\
        ContraWR        & $0.4678 \pm 0.0125$ & $0.2905 \pm 0.0160$ & $0.4413 \pm 0.0142$ & $0.5892 \pm 0.0133$ & $0.4527 \pm 0.0248$ & $0.5918 \pm 0.0116$ \\
        CNN-Transformer & $0.4600 \pm 0.0108$ & $0.2800 \pm 0.0148$ & $0.4460 \pm 0.0114$ & $0.6053 \pm 0.0118$ & $0.4725 \pm 0.0223$ & $0.6041 \pm 0.0105$ \\
        FFCL            & $0.4470 \pm 0.0143$ & $0.2627 \pm 0.0176$ & $0.4238 \pm 0.0139$ & $0.5726 \pm 0.0092$ & $0.4323 \pm 0.0182$ & $0.5701 \pm 0.0079$ \\
        ST-Transformer  & $0.4575 \pm 0.0145$ & $0.2733 \pm 0.0190$ & $0.4471 \pm 0.0142$ & $0.6035 \pm 0.0081$ & $0.4712 \pm 0.0199$ & $0.6053 \pm 0.0075$ \\
        \midrule
        BIOT            & $0.4748 \pm 0.0093$ & $0.2997 \pm 0.0134$ & $0.4607 \pm 0.0125$ & $0.6153 \pm 0.0154$ & $0.4875 \pm 0.0272$ & $0.6158 \pm 0.0197$ \\
        LaBraM-Base     & $0.4869 \pm 0.0085$ & $0.3159 \pm 0.0154$ & $0.4728 \pm 0.0105$ & $0.6173 \pm 0.0122$ & $0.4912 \pm 0.0192$ & $0.6177 \pm 0.0141$ \\
        CBraMod         & $0.5138 \pm 0.0066$ & $0.3518 \pm 0.0094$ & $0.4984 \pm 0.0085$ & $0.6417 \pm 0.0091$ & $0.5222 \pm 0.0169$ & $0.6427 \pm 0.0100$ \\
        REVE-Base       & $0.6396 \pm 0.0095$ & $0.5194 \pm 0.0126$ & $0.6339 \pm 0.0110$ & $0.6480 \pm 0.0140$ & $0.5306 \pm 0.0187$ & $0.6484 \pm 0.0170$ \\
        CSBrain         & $0.5657 \pm 0.0071$ & $0.4209 \pm 0.0093$ & $0.5637 \pm 0.0087$ & $0.6304 \pm 0.0090$ & $0.5071 \pm 0.0120$ & $0.6308 \pm 0.0095$ \\
        EEGMamba        & -- & -- & -- & $0.6323 \pm 0.0084$ & $0.5131 \pm 0.0124$ & $0.6342 \pm 0.0098$ \\
        \midrule
        \textbf{SPERA}   & \textbf{0.6816 $\pm$ 0.0147} & \textbf{0.5755 $\pm$ 0.0197} & \textbf{0.6817 $\pm$ 0.0155} & \textbf{0.6726 $\pm$ 0.0072} & \textbf{0.5633 $\pm$ 0.0096} & \textbf{0.6731 $\pm$ 0.0080} \\
        \bottomrule
    \end{tabular}
    }
\end{table}

\subsection{Emotion recognition}
FACED~\cite{chen2023large} is a 9-class emotion recognition dataset comprising 123 subjects recorded using 32 channels at 250 Hz, covering amusement, inspiration, joy, tenderness, anger, fear, disgust, sadness, and neutral. 
EEG signals are resampled to 200 Hz and segmented into 10-second samples, yielding 10,332 samples in total.
Subjects 1--80, 81--100, and 101--123 are used for training, validation, and test, respectively. 

Detailed results for emotion recognition are shown in Table~\ref{tab:faced}.
SPERA and CSBrain achieve nearly identical balanced accuracy (0.5754 vs. 0.5752). CodeBrain reports higher values on all three metrics; FACED is the only task among the nine where a recent model leads SPERA on every metric.

\begin{table}[htbp]
    \centering
    \caption{The results of different methods on emotion recognition (FACED, 9-class).}
    \label{tab:faced}
    % \vspace{2mm}
    \begin{tabular}{lccc}
        \toprule
        \textbf{Methods} & \textbf{Balanced accuracy} & \textbf{Cohen's $\kappa$} & \textbf{Weighted F1} \\
        \midrule
        EEGNet          & $0.4090 \pm 0.0122$ & $0.3342 \pm 0.0251$ & $0.4124 \pm 0.0141$ \\
        EEGConformer    & $0.4559 \pm 0.0125$ & $0.3858 \pm 0.0186$ & $0.4514 \pm 0.0107$ \\
        SPaRCNet        & $0.4673 \pm 0.0155$ & $0.3978 \pm 0.0289$ & $0.4729 \pm 0.0133$ \\
        ContraWR        & $0.4887 \pm 0.0078$ & $0.4231 \pm 0.0151$ & $0.4884 \pm 0.0074$ \\
        CNN-Transformer & $0.4697 \pm 0.0132$ & $0.4017 \pm 0.0168$ & $0.4720 \pm 0.0125$ \\
        FFCL            & $0.4673 \pm 0.0158$ & $0.3987 \pm 0.0383$ & $0.4699 \pm 0.0145$ \\
        ST-Transformer  & $0.4810 \pm 0.0079$ & $0.4137 \pm 0.0133$ & $0.4795 \pm 0.0096$ \\
        \midrule
        BIOT            & $0.5118 \pm 0.0118$ & $0.4476 \pm 0.0254$ & $0.5136 \pm 0.0112$ \\
        LaBraM-Base     & $0.5273 \pm 0.0107$ & $0.4698 \pm 0.0188$ & $0.5288 \pm 0.0102$ \\
        CBraMod         & $0.5509 \pm 0.0089$ & $0.5041 \pm 0.0122$ & $0.5618 \pm 0.0093$ \\
        REVE-Base       & $0.5646 \pm 0.0164$ & $0.5080 \pm 0.0191$ & $0.5659 \pm 0.0172$ \\
        CSBrain         & $0.5752 \pm 0.0042$ & $0.5204 \pm 0.0036$ & $0.5796 \pm 0.0031$ \\
        CodeBrain       & \textbf{0.5941 $\pm$ 0.0098} & \textbf{0.5406 $\pm$ 0.0084} & \textbf{0.5953 $\pm$ 0.0113} \\
        EEGMamba        & $0.5646 \pm 0.0096$ & $0.5134 \pm 0.0127$ & $0.5673 \pm 0.0076$ \\
        \midrule
        \textbf{SPERA}   & $0.5754 \pm 0.0175$ & $0.5122 \pm 0.0286$ & $0.5711 \pm 0.0138$ \\
        \bottomrule
    \end{tabular}
\end{table}

\subsection{Mental stress detection}
Mental arithmetic (MAT)~\cite{zyma2019electroencephalograms} is a 2-class mental stress dataset comprising 36 subjects recorded using 20 channels at 500 Hz, distinguishing resting state from active mental arithmetic. 
EEG signals are bandpass filtered (0.5--45 Hz), resampled to 200 Hz, and segmented into 1{,}707 5-second samples. 
Subjects 1--28, 29--32, and 33--36 are used for training, validation, and test, respectively. 

The experimental results are shown in Table~\ref{tab:mat}.
SPERA exceeds REVE-Base by $+$0.0118 in balanced accuracy and $+$0.0454 in AUROC, while its AUC-PR is lower. 

\begin{table}[ht]
    \centering
    \caption{The results of different methods on mental stress detection (MAT, 2-class).}
    \label{tab:mat}
    % \vspace{2mm}
    \begin{tabular}{lccc}
        \toprule
        \textbf{Methods} & \textbf{Balanced accuracy} & \textbf{AUC-PR} & \textbf{AUROC} \\
        \midrule
        EEGNet          & $0.6770 \pm 0.0116$ & $0.5763 \pm 0.0102$ & $0.7321 \pm 0.0108$ \\
        EEGConformer    & $0.6805 \pm 0.0123$ & $0.5829 \pm 0.0134$ & $0.7424 \pm 0.0128$ \\
        SPaRCNet        & $0.6879 \pm 0.0107$ & $0.5825 \pm 0.0193$ & $0.7418 \pm 0.0132$ \\
        ContraWR        & $0.6631 \pm 0.0097$ & $0.5787 \pm 0.0164$ & $0.7332 \pm 0.0082$ \\
        CNN-Transformer & $0.6779 \pm 0.0268$ & $0.5777 \pm 0.0285$ & $0.7258 \pm 0.0336$ \\
        FFCL            & $0.6798 \pm 0.0142$ & $0.5786 \pm 0.0266$ & $0.7330 \pm 0.0198$ \\
        ST-Transformer  & $0.6631 \pm 0.0173$ & $0.5672 \pm 0.0259$ & $0.7132 \pm 0.0174$ \\
        \midrule
        BIOT            & $0.6875 \pm 0.0186$ & $0.6004 \pm 0.0195$ & $0.7536 \pm 0.0144$ \\
        LaBraM-Base     & $0.6909 \pm 0.0125$ & $0.5999 \pm 0.0155$ & $0.7721 \pm 0.0093$ \\
        CBraMod         & $0.7256 \pm 0.0132$ & $0.6267 \pm 0.0099$ & $0.7905 \pm 0.0073$ \\
        REVE-Base       & $0.7660 \pm 0.0355$ & \textbf{0.7470 $\pm$ 0.0807} & $0.8450 \pm 0.0514$ \\
        CSBrain         & $0.7558 \pm 0.0106$ & $0.6696 \pm 0.0221$ & $0.8478 \pm 0.0297$ \\
        CodeBrain       & $0.7514 \pm 0.0203$ & $0.7177 \pm 0.0421$ & $0.8707 \pm 0.0209$ \\
        \midrule
        \textbf{SPERA}   & \textbf{0.7778 $\pm$ 0.0216} & 0.7291 $\pm$ 0.0226 & \textbf{0.8904 $\pm$ 0.0196} \\
        \bottomrule
    \end{tabular}
\end{table}

\subsection{Sleep staging}

\begin{table}[t]
    \centering
    \caption{The results of different methods on sleep staging (ISRUC, 5-class).}
    \label{tab:isruc}
    % \vspace{2mm}
    \begin{tabular}{lccc}
        \toprule
        \textbf{Methods} & \textbf{Balanced accuracy} & \textbf{Cohen's $\kappa$} & \textbf{Weighted F1} \\
        \midrule
        EEGNet          & $0.7154 \pm 0.0121$ & $0.7040 \pm 0.0173$ & $0.7513 \pm 0.0124$ \\
        EEGConformer    & $0.7400 \pm 0.0133$ & $0.7143 \pm 0.0162$ & $0.7634 \pm 0.0150$ \\
        SPaRCNet        & $0.7487 \pm 0.0075$ & $0.7097 \pm 0.0132$ & $0.7624 \pm 0.0092$ \\
        ContraWR        & $0.7402 \pm 0.0126$ & $0.7178 \pm 0.0156$ & $0.7610 \pm 0.0137$ \\
        CNN-Transformer & $0.7363 \pm 0.0087$ & $0.7129 \pm 0.0121$ & $0.7719 \pm 0.0105$ \\
        FFCL            & $0.7277 \pm 0.0182$ & $0.7016 \pm 0.0291$ & $0.7614 \pm 0.0197$ \\
        ST-Transformer  & $0.7381 \pm 0.0205$ & $0.7013 \pm 0.0352$ & $0.7681 \pm 0.0175$ \\
        DeepSleepNet    & $0.7419 \pm 0.0144$ & $0.7036 \pm 0.0241$ & $0.7643 \pm 0.0122$ \\
        U-Sleep         & $0.7586 \pm 0.0116$ & $0.7209 \pm 0.0143$ & $0.7805 \pm 0.0105$ \\
        \midrule
        BIOT            & $0.7527 \pm 0.0121$ & $0.7192 \pm 0.0231$ & $0.7790 \pm 0.0146$ \\
        LaBraM-Base     & $0.7633 \pm 0.0102$ & $0.7231 \pm 0.0182$ & $0.7810 \pm 0.0133$ \\
        CBraMod         & $0.7865 \pm 0.0110$ & $0.7442 \pm 0.0152$ & $0.8011 \pm 0.0099$ \\
        REVE-Base       & $0.7819 \pm 0.0078$ & $0.7500 \pm 0.0156$ & $0.8005 \pm 0.0135$ \\
        CSBrain         & $0.7925 \pm 0.0030$ & $0.7406 \pm 0.0102$ & $0.7990 \pm 0.0091$ \\
        CodeBrain       & $0.7835 \pm 0.0033$ & $0.7476 \pm 0.0040$ & $0.8020 \pm 0.0018$ \\
        EEGMamba        & $0.7899 \pm 0.0093$ & $0.7486 \pm 0.0122$ & $0.8054 \pm 0.0076$ \\
        \midrule
        \textbf{SPERA (single-segment)}   & 0.8093 $\pm$ 0.0108 & 0.7542 $\pm$ 0.0187 & 0.8115 $\pm$ 0.0150 \\
        \textbf{SPERA (sequence-to-sequence)} & \textbf{0.8137 $\pm$ 0.0075} & \textbf{0.7662 $\pm$ 0.0121} & \textbf{0.8217 $\pm$ 0.0097} \\
        \bottomrule
    \end{tabular}
\end{table}

ISRUC~\cite{khalighi2016isruc} is a 5-class sleep staging dataset comprising 100 subjects recorded using 6 channels at 200 Hz. 
EEG signals are segmented into 30-second epochs and annotated according to the AASM standard (Wake, N1, N2, N3, REM). 
Subjects 1--80, 81--90, and 91--100 are used for training, validation, and test, respectively.
Subject 8 was excluded from the final dataset due to missing requisite EEG channels (e.g., F3, F4), resulting in a total of 88,140 samples.
Sleep staging is commonly framed as a sequence-to-sequence classification problem with a sequence length of 20 consecutive 30-second epochs to model sleep stage transitions~\cite{phan2022automatic}. 
Following the protocol of CBraMod~\cite{wang2024cbramod}, we adopt this standard setting and additionally report single-segment classification results to assess the quality of our pretrained representations without sequential context.
We further include two widely recognized sleep-staging-specific baselines, DeepSleepNet~\cite{supratak2017deepsleepnet} and U-Sleep~\cite{perslev2021u}, in order to compare against classical methods established in the sleep staging literature. 

Full classification results are summarized in Table~\ref{tab:isruc}.
The single-segment variant alone (0.8093) exceeds the strongest baseline (CSBrain, 0.7925) in balanced accuracy, indicating that the gain does not require the sequence-to-sequence formulation. The sequence-to-sequence variant further improves performance to 0.8137. 
These results indicate that SPERA remains effective on ISRUC, which has the fewest channels (6) and the longest segment duration (30 s) among the nine downstream datasets.

\subsection{Mental disorder diagnosis}
Mumtaz~\cite{Mumtaz2016} is a 2-class depression detection dataset comprising 34 patients with major depressive disorder (MDD) and 30 normal controls, recorded using 19 channels at 256 Hz. 
Only eyes-open and eyes-closed resting-state sessions are used; MDD subject 8 is excluded because only a P300 recording is available. 
EEG signals are bandpass filtered (0.3--75 Hz), notch filtered (50 Hz), resampled to 200 Hz, and segmented into 5-second samples. 
Following the publicly released preprocessing implementation of prior EEG foundation model studies~\cite{wang2024cbramod, ouahidi2025reve}\footnote{The textual descriptions in~\cite{wang2024cbramod, ouahidi2025reve} report subject counts of 24/5/5 MDD and 19/4/5 controls, whereas applying the released preprocessing code to the resting-state recordings yields 22/5/6 and 21/4/5, respectively. We follow the released implementation, which determines the data partition used in the experiments.}, we apply the released train/validation/test partition. 
The resulting split contains 22/5/6 MDD and 21/4/5 control subjects (33 MDD and 30 controls in total), yielding a total of 7,143 5-second samples consistent with prior work.

The experimental results are summarized in Table~\ref{tab:mumtaz}.
SPERA and REVE-Base are comparable in balanced accuracy (0.9637 vs.\ 0.9644), with SPERA achieving the highest AUC-PR and AUROC. 
SPERA also exhibits the smallest across-seed standard deviation in balanced accuracy ($\pm$ 0.0018) among all compared methods.

\begin{table}[t]
    \centering
    \caption{The results of different methods on mental disorder diagnosis (Mumtaz, 2-class).}
    \label{tab:mumtaz}
    % \vspace{2mm}
    \begin{tabular}{lccc}
        \toprule
        \textbf{Methods} & \textbf{Balanced accuracy} & \textbf{AUC-PR} & \textbf{AUROC} \\
        \midrule
        EEGNet          & $0.9232 \pm 0.0104$ & $0.9626 \pm 0.0095$ & $0.9639 \pm 0.0093$ \\
        EEGConformer    & $0.9308 \pm 0.0117$ & $0.9684 \pm 0.0105$ & $0.9702 \pm 0.0101$ \\
        SPaRCNet        & $0.9316 \pm 0.0095$ & $0.9754 \pm 0.0065$ & $0.9781 \pm 0.0083$ \\
        ContraWR        & $0.9195 \pm 0.0115$ & $0.9589 \pm 0.0102$ & $0.9621 \pm 0.0092$ \\
        CNN-Transformer & $0.9305 \pm 0.0068$ & $0.9757 \pm 0.0074$ & $0.9742 \pm 0.0055$ \\
        FFCL            & $0.9314 \pm 0.0038$ & $0.9717 \pm 0.0021$ & $0.9753 \pm 0.0033$ \\
        ST-Transformer  & $0.9135 \pm 0.0103$ & $0.9578 \pm 0.0086$ & $0.9594 \pm 0.0059$ \\
        \midrule
        BIOT            & $0.9358 \pm 0.0052$ & $0.9736 \pm 0.0034$ & $0.9758 \pm 0.0042$ \\
        LaBraM-Base     & $0.9409 \pm 0.0079$ & $0.9798 \pm 0.0093$ & $0.9782 \pm 0.0057$ \\
        CBraMod         & $0.9560 \pm 0.0056$ & $0.9923 \pm 0.0032$ & $0.9921 \pm 0.0025$ \\
        REVE-Base       & \textbf{0.9644 $\pm$ 0.0097} & $0.9961 \pm 0.0013$ & $0.9957 \pm 0.0015$ \\
        CSBrain         & $0.9643 \pm 0.0155$ & $0.9936 \pm 0.0034$ & $0.9956 \pm 0.0020$ \\
        \midrule
        \textbf{SPERA}   & 0.9637 $\pm$ 0.0018 & \textbf{0.9973 $\pm$ 0.0007} & \textbf{0.9975 $\pm$ 0.0010} \\
        \bottomrule
    \end{tabular}
\end{table}

\subsection{EEG abnormality detection}
TUAB~\cite{obeid2016temple} is a 2-class EEG abnormality detection dataset recorded using 23 channels at 250 Hz, with 16 common bipolar montage channels used for analysis.
EEG signals are resampled to 200 Hz and segmented into 409,455 10-second samples. 
The original train/test split is adopted, with the training set further divided into training and validation subsets (80\%:20\%). 
TUAB is excluded from our pretraining corpus, preventing any evaluation leakage.

\begin{table}[ht]
    \centering
    \caption{The results of different methods on EEG abnormality detection (TUAB, 2-class).}
    \label{tab:tuab}
    % \vspace{2mm}
    \begin{tabular}{lccc}
        \toprule
        \textbf{Methods} & \textbf{Balanced accuracy} & \textbf{AUC-PR} & \textbf{AUROC} \\
        \midrule
        EEGNet          & $0.7642 \pm 0.0036$ & $0.8299 \pm 0.0043$ & $0.8412 \pm 0.0030$ \\
        EEGConformer    & $0.7758 \pm 0.0049$ & $0.8427 \pm 0.0054$ & $0.8445 \pm 0.0038$ \\
        SPaRCNet        & $0.7896 \pm 0.0018$ & $0.8414 \pm 0.0018$ & $0.8676 \pm 0.0011$ \\
        ContraWR        & $0.7746 \pm 0.0041$ & $0.8421 \pm 0.0104$ & $0.8456 \pm 0.0074$ \\
        CNN-Transformer & $0.7777 \pm 0.0022$ & $0.8433 \pm 0.0039$ & $0.8461 \pm 0.0013$ \\
        FFCL            & $0.7848 \pm 0.0038$ & $0.8448 \pm 0.0065$ & $0.8569 \pm 0.0051$ \\
        ST-Transformer  & $0.7966 \pm 0.0023$ & $0.8521 \pm 0.0026$ & $0.8707 \pm 0.0019$ \\
        EEG2Rep         & -- & -- & $0.8843 \pm 0.0309$ \\
        \midrule
        BIOT            & $0.7959 \pm 0.0057$ & $0.8792 \pm 0.0023$ & $0.8815 \pm 0.0043$ \\
        LaBraM-Base     & $0.8140 \pm 0.0019$ & $0.8965 \pm 0.0016$ & $0.9022 \pm 0.0008$ \\
        LaBraM-Large    & $0.8226 \pm 0.0015$ & $0.9130 \pm 0.0005$ & $0.9127 \pm 0.0005$ \\
        LaBraM-Huge     & $0.8258 \pm 0.0011$ & $0.9204 \pm 0.0011$ & $0.9162 \pm 0.0016$ \\
        CBraMod         & $0.8289 \pm 0.0022$ & $0.9258 \pm 0.0008$ & $0.9227 \pm 0.0011$ \\
        REVE-Base       & $0.8315 \pm 0.0014$ & $0.9281 \pm 0.0009$ & $0.9245 \pm 0.0013$ \\
        CSBrain         & $0.8172 \pm 0.0043$ & $0.9005 \pm 0.0066$ & $0.8957 \pm 0.0046$ \\
        CodeBrain       & $0.8294 \pm 0.0013$ & $0.9100 \pm 0.0006$ & $0.9030 \pm 0.0009$ \\
        LuMamba         & $0.8099 \pm 0.0022$ & $0.8918 \pm 0.0032$ & $0.8825 \pm 0.0038$ \\
        \midrule
        \textbf{SPERA}   & \textbf{0.8488 $\pm$ 0.0035} & \textbf{0.9287 $\pm$ 0.0045} & \textbf{0.9253 $\pm$ 0.0046} \\
        \bottomrule
    \end{tabular}
\end{table}

As shown in Table~\ref{tab:tuab}, SPERA improves balanced accuracy by $+$0.0173 over REVE-Base, with smaller gains in AUC-PR and AUROC. 

\subsection{Event type classification}

TUEV~\cite{obeid2016temple} is a 6-class EEG event classification dataset recorded using 23 channels at 250 Hz, covering spike and sharp wave, generalized periodic epileptiform discharges, periodic lateralized epileptiform discharges, eye movement, artifact, and background. 
EEG signals from 16 common bipolar montage channels are resampled to 200 Hz and segmented into 112,491 5-second samples. 
The original train/test split is adopted, with the training set further divided into training and validation subsets (80\%:20\%). 
As with TUAB, TUEV is excluded from our pretraining corpus. 

The experimental results are summarized in Table~\ref{tab:tuev}.
SPERA achieves the highest balanced accuracy among the evaluated methods (0.7684). EEGDM reports higher Cohen's $\kappa$ and weighted F1 than SPERA, while GRAM reports the highest weighted F1.

\begin{table}[ht]
    \centering
    \caption{The results of different methods on event type classification (TUEV, 6-class).}
    \label{tab:tuev}
    % \vspace{2mm}
    \begin{tabular}{lccc}
        \toprule
        \textbf{Methods} & \textbf{Balanced accuracy} & \textbf{Cohen's $\kappa$} & \textbf{Weighted F1} \\
        \midrule
        EEGNet          & $0.3876 \pm 0.0143$ & $0.3577 \pm 0.0155$ & $0.6539 \pm 0.0120$ \\
        EEGConformer    & $0.4074 \pm 0.0164$ & $0.3967 \pm 0.0195$ & $0.6983 \pm 0.0152$ \\
        SPaRCNet        & $0.4161 \pm 0.0262$ & $0.4233 \pm 0.0181$ & $0.7024 \pm 0.0104$ \\
        ContraWR        & $0.4384 \pm 0.0349$ & $0.3912 \pm 0.0237$ & $0.6893 \pm 0.0136$ \\
        CNN-Transformer & $0.4087 \pm 0.0161$ & $0.3815 \pm 0.0134$ & $0.6854 \pm 0.0290$ \\
        FFCL            & $0.3979 \pm 0.0104$ & $0.3732 \pm 0.0188$ & $0.6783 \pm 0.0120$ \\
        ST-Transformer  & $0.3984 \pm 0.0228$ & $0.3765 \pm 0.0306$ & $0.6823 \pm 0.0190$ \\
        EEG2Rep         & $0.5295 \pm 0.0158$ & -- & $0.7508 \pm 0.0121$ \\
        EEGDM           & $0.7557 \pm 0.0147$ & \textbf{0.7423 $\pm$ 0.0136} & $0.8688 \pm 0.0066$ \\
        \midrule
        BIOT            & $0.5281 \pm 0.0225$ & $0.5273 \pm 0.0249$ & $0.7492 \pm 0.0080$ \\
        LaBraM-Base     & $0.6409 \pm 0.0065$ & $0.6637 \pm 0.0093$ & $0.8312 \pm 0.0052$ \\
        LaBraM-Large    & $0.6581 \pm 0.0156$ & $0.6622 \pm 0.0136$ & $0.8315 \pm 0.0040$ \\
        LaBraM-Huge     & $0.6616 \pm 0.0170$ & $0.6745 \pm 0.0195$ & $0.8329 \pm 0.0086$ \\
        CBraMod         & $0.6671 \pm 0.0107$ & $0.6772 \pm 0.0096$ & $0.8342 \pm 0.0064$ \\
        REVE-Base       & $0.6759 \pm 0.0229$ & $0.6783 \pm 0.0199$ & $0.8451 \pm 0.0129$ \\
        CSBrain         & $0.6903 \pm 0.0059$ & $0.6833 \pm 0.0047$ & $0.8333 \pm 0.0057$ \\
        CodeBrain       & $0.6428 \pm 0.0062$ & $0.6912 \pm 0.0101$ & $0.8362 \pm 0.0048$ \\
        GRAM            & $0.7487 \pm 0.0151$ & $0.7130 \pm 0.0333$ & \textbf{0.8824 $\pm$ 0.0158} \\
        \midrule
        \textbf{SPERA}   & \textbf{0.7684 $\pm$ 0.0283} & $0.7320 \pm 0.0366$ & $0.8619 \pm 0.0178$ \\
        \bottomrule
    \end{tabular}
\end{table}

\subsection{Linear probing results}
\label{app:lp}

Table~\ref{tab:lp} reports the full per-task LP results across all nine downstream datasets. 
The largest gains over baselines are observed on BCIC2020-3 and TUEV, where SPERA-Large surpasses REVE-Large by over 0.1 in balanced accuracy.
On BCIC-IV-2a, PhysioNet-MI, MAT, and Mumtaz, SPERA models reach competitive but not superior performance. Mumtaz in particular shows limited headroom, with most foundation models exceeding 0.9 balanced accuracy.
While most tasks improve with model size, PhysioNet-MI, FACED, and MAT show non-monotonic scaling under LP.

\begin{table}[h]
    \centering
    \captionsetup{font=normalsize}
    \caption{LP results on nine downstream tasks. The results are reported as mean balanced accuracy $\pm$ standard deviation over five random seeds. The best performance for each task is highlighted in bold. Baseline model results are extracted from~\cite{ouahidi2025reve}. The standard deviations for the average performance of the baseline models are omitted as they are not reported in the original paper.}
    % \vspace{2mm}
    \label{tab:lp}
    \resizebox{\textwidth}{!}{
    \begin{tabular}{lcccccc}
        \toprule
        \textbf{Methods} & \textbf{Params} & \textbf{BCIC2020-3} & \textbf{BCIC-IV-2a} & \textbf{PhysioNet-MI} & \textbf{FACED} & \textbf{MAT} \\
        \midrule
        CBraMod & 4M & $0.374 \pm 0.007$ & $0.376 \pm 0.006$ & $0.531 \pm 0.015$ & $0.261 \pm 0.013$ & $0.605 \pm 0.020$ \\
        REVE-Base & 69M & $0.390 \pm 0.017$ & $0.517 \pm 0.015$ & $0.510 \pm 0.012$ & $0.422 \pm 0.028$ & \textbf{0.740 $\pm$ 0.073} \\
        REVE-Large & 408M & $0.378 \pm 0.021$ & \textbf{0.603 $\pm$ 0.011} & \textbf{0.617 $\pm$ 0.000} & $0.469 \pm 0.007$ & $0.665 \pm 0.103$ \\
        \midrule
        SPERA-Tiny & 18M & $0.324 \pm 0.024$ & $0.497 \pm 0.009$ & $0.532 \pm 0.007$ & $0.343 \pm 0.009$ & $0.689 \pm 0.014$ \\
        SPERA-Small & 48M & $0.407 \pm 0.015$ & $0.497 \pm 0.013$ & $0.523 \pm 0.011$ & $0.363 \pm 0.017$ & $0.630 \pm 0.022$ \\
        SPERA-Base & 194M & 0.502 $\pm$ 0.011 & 0.560 $\pm$ 0.010 & 0.551 $\pm$ 0.009 & \textbf{0.478 $\pm$ 0.011} & 0.672 $\pm$ 0.014\\
        SPERA-Large & 396M & \textbf{0.593 $\pm$ 0.019} & 0.590 $\pm$ 0.014 & 0.583 $\pm$ 0.008 & 0.469 $\pm$ 0.010 & 0.693 $\pm$ 0.016 \\
        \midrule \midrule
        
        \textbf{Methods} & \textbf{Params} & \textbf{ISRUC} & \textbf{Mumtaz} & \textbf{TUAB} & \textbf{TUEV} & \textbf{Average} \\
        \midrule
        CBraMod & 4M & $0.430 \pm 0.043$ & $0.907 \pm 0.027$ & $0.500 \pm 0.000$ & $0.482 \pm 0.037$ & 0.496 \\
        REVE-Base & 69M & $0.662 \pm 0.030$ & $0.931 \pm 0.021$ & $0.809 \pm 0.004$ & $0.508 \pm 0.073$ & 0.610 \\
        REVE-Large & 408M & $0.758 \pm 0.001$ & \textbf{0.980 $\pm$ 0.009} & $0.809 \pm 0.004$ & $0.550 \pm 0.014$ & 0.648 \\
        \midrule
        SPERA-Tiny & 18M & $0.687 \pm 0.027$ & $0.885 \pm 0.029$ & $0.784 \pm 0.003$ & $0.517 \pm 0.007$ & $0.584 \pm 0.006$ \\
        SPERA-Small & 48M & $0.721 \pm 0.024$ & $0.929 \pm 0.011$ & $0.787 \pm 0.006$ & $0.558 \pm 0.010$ & $0.601 \pm 0.005$ \\
        SPERA-Base & 194M & 0.772 $\pm$ 0.014 & 0.944 $\pm$ 0.015 & 0.821 $\pm$ 0.002 & 0.651 $\pm$ 0.008 & 0.661 $\pm$ 0.003\\
        SPERA-Large & 396M & \textbf{0.775 $\pm$ 0.019} & 0.957 $\pm$ 0.032 & \textbf{0.842 $\pm$ 0.003} & \textbf{0.653 $\pm$ 0.008} & \textbf{0.684 $\pm$ 0.005} \\
        \bottomrule
    \end{tabular}
    }
\end{table}

Table~\ref{tab:lp_recent} additionally compares SPERA-Base under LP with recent models on TUAB and TUEV, the tasks from our evaluation suite that overlap with their reported evaluations.
SPERA-Base attains the highest balanced accuracy on both tasks.
On TUEV, however, EEGPT reports a higher Cohen's $\kappa$ and weighted F1.

\begin{table}[h]
    \centering
    \captionsetup{font=normalsize}
    \caption{LP comparison with recent models on TUAB and TUEV. For the binary TUAB task, we report AUC-PR and AUROC; for the 6-class TUEV task, we report Cohen's $\kappa$ and weighted F1. Baseline numbers are taken from the original publications. Laya evaluates TUAB under its own LP protocol, in which reproduced baseline scores differ substantially from those in Table~\ref{tab:lp}.}
    \label{tab:lp_recent}
    \resizebox{\textwidth}{!}{
    \begin{tabular}{lcccccc}
        \toprule
        \multirow{2}{*}{\textbf{Methods}} & \multicolumn{3}{c}{\textbf{TUAB, 2-class}} & \multicolumn{3}{c}{\textbf{TUEV, 6-class}} \\
        \cmidrule(lr){2-4} \cmidrule(lr){5-7}
        & \textbf{Balanced accuracy} & \textbf{AUC-PR} & \textbf{AUROC} & \textbf{Balanced accuracy} & \textbf{Cohen's $\kappa$} & \textbf{Weighted F1} \\
        \midrule
        EEGPT       & $0.798 \pm 0.003$ & -- & $0.872 \pm 0.005$ & $0.623 \pm 0.011$ & \textbf{0.635 $\pm$ 0.013} & \textbf{0.819 $\pm$ 0.006} \\
        EEG2Rep     & -- & -- & $0.832 \pm 0.033$ & $0.433 \pm 0.031$ & -- & $0.700 \pm 0.032$ \\
        Laya        & $0.798 \pm 0.007$ & -- & -- & -- & -- & -- \\
        \midrule
        \textbf{SPERA-Base} & \textbf{0.821 $\pm$ 0.002} & \textbf{0.898 $\pm$ 0.001} & \textbf{0.901 $\pm$ 0.002} & \textbf{0.651 $\pm$ 0.008} & $0.580 \pm 0.015$ & $0.784 \pm 0.008$ \\
        \bottomrule
    \end{tabular}
    }
\end{table}

\begin{table}[t]
\centering
\caption{Per-task significance under LPFT. The comparator is the highest-balanced-accuracy baseline among methods reporting results on all nine downstream tasks; methods with partial task coverage are excluded. $\Delta$ denotes the balanced-accuracy difference relative to the comparator, brackets show 95\% confidence intervals, and bold indicates $p_{\mathrm{Holm}}<0.05$.}
\label{app:tab:significance}
\resizebox{\textwidth}{!}{
\begin{tabular}{llcccccc}
\toprule
& & \multicolumn{3}{c}{\textbf{SPERA-Base (194M)}} & \multicolumn{3}{c}{\textbf{SPERA-Large (396M)}} \\
\cmidrule(lr){3-5}\cmidrule(lr){6-8}
\textbf{Task} & \textbf{Comparator} & $\boldsymbol{\Delta}$ \textbf{[95\% CI]} & $\boldsymbol{p}$ & $\boldsymbol{p_{\text{Holm}}}$ & $\boldsymbol{\Delta}$ \textbf{[95\% CI]} & $\boldsymbol{p}$ & $\boldsymbol{p_{\text{Holm}}}$ \\
\midrule
BCIC2020-3 & CSBrain & $+0.0500$ [$+0.026$, $+0.074$] & 0.002 & \textbf{0.011} & $+0.0879$ [$+0.063$, $+0.112$] & $<0.001$ & \textbf{< 0.001} \\
TUEV & CSBrain & $+0.0781$ [$+0.043$, $+0.113$] & 0.003 & \textbf{0.014} & $+0.0966$ [$+0.061$, $+0.132$] & 0.001 & \textbf{0.008} \\
BCIC-IV-2a & REVE-Base & $+0.0420$ [$+0.023$, $+0.061$] & 0.001 & \textbf{0.009} & $+0.0627$ [$+0.044$, $+0.081$] & $<0.001$ & \textbf{< 0.001} \\
ISRUC & CSBrain & $+0.0212$ [$+0.012$, $+0.030$] & 0.002 & \textbf{0.011} & $+0.0217$ [$+0.009$, $+0.035$] & 0.008 & \textbf{0.033} \\
TUAB & REVE-Base & $+0.0173$ [$+0.013$, $+0.022$] & $<0.001$ & \textbf{0.001} & $+0.0293$ [$+0.026$, $+0.033$] & $<0.001$ & \textbf{< 0.001} \\
\midrule
PhysioNet-MI & REVE-Base & $+0.0246$ [$+0.007$, $+0.042$] & 0.013 & 0.052 & $+0.0337$ [$+0.016$, $+0.051$] & 0.003 & \textbf{0.014} \\
MAT & REVE-Base & $+0.0118$ [$-0.033$, $+0.056$] & 0.547 & 1.000 & $+0.0361$ [$-0.008$, $+0.080$] & 0.093 & 0.279 \\
FACED & CSBrain & $+0.0002$ [$-0.021$, $+0.022$] & 0.981 & 1.000 & $+0.0167$ [$-0.006$, $+0.040$] & 0.115 & 0.279 \\
Mumtaz & REVE-Base & $-0.0007$ [$-0.013$, $+0.011$] & 0.881 & 1.000 & $+0.0058$ [$-0.006$, $+0.018$] & 0.254 & 0.279 \\
\bottomrule
\end{tabular}}
\end{table}

\subsection{Per-task significance}
\label{app:significance}

We evaluate the statistical significance of SPERA's task-level performance differences across the nine downstream tasks.
For consistency across tasks, the comparator pool is restricted to baselines with balanced-accuracy results reported on all nine tasks (Table~\ref{tab:main}); methods with partial task coverage are excluded.
For each task, SPERA is compared with the highest-performing baseline in this pool using Welch's $t$-test over five random seeds, with Holm correction applied across the nine tasks separately for each SPERA model.
The tests are two-sided, and the 95\% confidence intervals are pointwise and are not adjusted for multiple comparisons.
The tests are unpaired because baseline results are taken from the original publications and per-seed values are unavailable.

Table~\ref{app:tab:significance} reports the difference in balanced accuracy $\Delta$ with its 95\% confidence interval, alongside raw and Holm-corrected $p$-values.
For SPERA-Base, five tasks remain significant after correction: BCIC2020-3, TUEV, BCIC-IV-2a, ISRUC, and TUAB. PhysioNet-MI is borderline ($p_{\mathrm{Holm}} =$ 0.052). The differences on FACED, MAT, and Mumtaz did not reach statistical significance after correction. 
Mumtaz shows limited headroom, with generally high balanced accuracy across foundation models and model scales.
For MAT, the estimated performance difference is small relative to the across-seed variability, and the confidence interval includes zero, whereas on FACED, SPERA-Base and CSBrain achieve nearly identical balanced accuracy (0.5754 vs. 0.5752).

SPERA-Large shows a larger $\Delta$ than SPERA-Base on every task: the same five tasks remain significant, and PhysioNet-MI also reaches significance after correction. FACED, MAT, and Mumtaz remain non-significant. We report the main-text claim for SPERA-Base, the default model used throughout the paper, and include SPERA-Large here for completeness.

\section{Additional ablation experiments}
\label{app:ablations}

\subsection{Pretraining objective and corpus}
\label{app:ablation:objective}
 
The ablations in Table~\ref{tab:ablation} vary architectural components while holding the primary prediction objective and corpus fixed.
Here we vary those two factors instead, to separate the contribution of latent-space prediction from that of the pretraining data. 
Both variants use the SPERA-Small backbone and follow the ablation protocol of Section~\ref{sec:ablation}: LP on PhysioNet-MI, ISRUC, and TUAB.
Each configuration is evaluated over five random seeds.
 
For the reconstruction variant, we replace the latent prediction objective with input-space reconstruction. 
The predictor outputs are decoded back to raw patches by a linear head and the model is trained against the masked input patches, so that the target is computed from the raw signal. 
All other components, including the Legendre prior, the hybrid backbone, and the relational spectral regularizer, are unchanged.

For the REVE corpus variant, we restrict pretraining to dataset sources shared with REVE~\cite{ouahidi2025reve} to assess the effect of this corpus restriction on SPERA's downstream performance.
The proxy comprises 81 datasets from our corpus that are also listed in REVE's pretraining corpus, totaling 61,191~h from 22,994 subjects, comparable in duration to the approximately 60,000~h reported for REVE.
The 25 excluded datasets account for 18,473~h and contribute 6,054 additional subjects; 16,367~h are sleep recordings from MESA and the HSP subset, and the remaining 2,106~h come from 23 datasets.
 
Replacing latent prediction with input-space reconstruction lowers average balanced accuracy by 0.022 with the backbone, corpus, and token budget held fixed, and the drop is consistent across all three tasks. 
Pretraining on the REVE corpus proxy lowers the average by 0.005.
Specifically, PhysioNet-MI and TUAB are nearly identical, while ISRUC is lower by 0.016.
This is consistent with the composition of the excluded data, 89\% of which is sleep recordings, and ISRUC is the only sleep-staging task among the three.
 
\begin{table}[ht]
\centering
\caption{Ablation of the pretraining objective and corpus. All variants use the SPERA-Small backbone under the ablation protocol of Section~\ref{sec:ablation}. Results are mean balanced accuracy $\pm$ standard deviation over five random seeds. \textit{Reconstruction head} replaces latent prediction with input-space reconstruction;
\textit{REVE corpus} pretrains on the 81-dataset proxy for the corpus used by REVE~\cite{ouahidi2025reve}.}
\label{app:tab:objective}
\resizebox{\textwidth}{!}{
\begin{tabular}{lcccc}
\toprule
\textbf{Variants} & \textbf{PhysioNet-MI} & \textbf{ISRUC} & \textbf{TUAB} & \textbf{Average} \\
\midrule
SPERA-Small (latent prediction) & \textbf{0.523 $\pm$ 0.011} & \textbf{0.721 $\pm$ 0.024} & $0.787 \pm 0.006$ & \textbf{0.677 $\pm$ 0.011} \\
\quad with reconstruction head    & $0.499 \pm 0.015$ & $0.706 \pm 0.019$ & $0.761 \pm 0.011$ & $0.655 \pm 0.013$ \\
\quad with REVE corpus            & $0.522 \pm 0.013$ & $0.705 \pm 0.021$ & \textbf{0.788 $\pm$ 0.008} & $0.672 \pm 0.010$ \\
\bottomrule
\end{tabular}}
\end{table}

\subsection{Temporal positional encoding}
\label{app:ablation:pe}
 
SPERA applies rotary position embeddings (RoPE) along the temporal patch axis, while electrode position is supplied by the Legendre prior along the spatial axis. 
We ablate the temporal encoding by replacing RoPE with a standard sinusoidal absolute position embedding added to the patch embeddings, keeping the spatial encoding, the backbone, and the pretraining objective fixed. 
The variant follows the ablation protocol of Section~\ref{sec:ablation}. 
Note that the substitution affects the temporal axis only: in the factorized spatial pass RoPE is already disabled, and the geometry encodings continue to provide positional information there.
 
The result is summarized in Table~\ref{app:tab:temporal_pe}. 
RoPE is stronger on all three tasks, by 0.022 on average, supporting its use as the temporal positional encoding in the present architecture. This result is also consistent with prior reports that relative positional schemes can generalize across varying sequence lengths~\cite{su2024roformer}. 
The three tasks in this ablation use 4-s, 30-s, and 10-s segments for PhysioNet-MI, ISRUC, and TUAB, respectively, corresponding to 4, 30, and 10 temporal tokens per channel with 1-s patches.
 
\begin{table}[t]
\centering
\caption{Ablation of the temporal positional encoding. Both variants use the SPERA-Small backbone
under the ablation protocol of Section~\ref{sec:ablation}. Results are mean balanced accuracy $\pm$ standard deviation over five random seeds; the best result in each column is in bold. The spatial encoding is unchanged in
both variants.}
\label{app:tab:temporal_pe}
\resizebox{\textwidth}{!}{
\begin{tabular}{lcccc}
\toprule
\textbf{Variants} & \textbf{PhysioNet-MI} & \textbf{ISRUC} & \textbf{TUAB} & \textbf{Average} \\
\midrule
RoPE (SPERA-Small)   & \textbf{0.523 $\pm$ 0.011} & \textbf{0.721 $\pm$ 0.024} & \textbf{0.787 $\pm$ 0.006} & \textbf{0.677 $\pm$ 0.011} \\
Sinusoidal absolute encoding & $0.502 \pm 0.013$ & $0.699 \pm 0.020$ & $0.764 \pm 0.010$ & $0.655 \pm 0.011$ \\
\bottomrule
\end{tabular}}
\end{table}

\subsection{Masking strategy}
\label{app:masking}

To assess the impact of our masking strategy, we conduct an ablation comparing two strategies: block masking and semantic subsequence preserving (SSP) masking. 
We evaluate both strategies across three downstream tasks at five levels of expected masking ratio under LP using SPERA-Small.
The total masking ratio is determined by independent temporal and spatial masking processes, each applied with probability $p=$ 0.8. 
To prevent the trivial case where no tokens are masked, temporal masking is applied as a fallback.
The expected total masking ratio $\rho$ is computed as the expected union of the masked regions, assuming independence between temporal and spatial masking. We define five difficulty levels ($\rho \approx$ 0.27, 0.34, 0.45, 0.65, 0.75); detailed hyperparameter ranges for each level are summarized in Table~\ref{tab:masking_params}.

\begin{table}[h]
\centering
\caption{Masking ratio hyperparameters per difficulty level. The expected masking ratio $\rho$ accounts for the combined application of spatial and temporal masking, including the fallback condition.}
% \vspace{2mm}
\label{tab:masking_params}
    \resizebox{\textwidth}{!}{
    \begin{tabular}{lccccc}
    \toprule
    \textbf{Level} & \textbf{Spatial range} & \textbf{Spatial mean} & \textbf{Temporal range} & \textbf{Temporal mean} & \textbf{Expected ratio ($\rho$)} \\
    \midrule
    0 & 0.05 -- 0.25 & 0.15 & 0.10 -- 0.30 & 0.20 & \textbf{0.27} \\
    1 & 0.10 -- 0.30 & 0.20 & 0.15 -- 0.35 & 0.25 & \textbf{0.34} \\
    2 & 0.18 -- 0.32 & 0.25 & 0.30 -- 0.44 & 0.37 & \textbf{0.45} \\
    3 & 0.27 -- 0.43 & 0.35 & 0.52 -- 0.68 & 0.60 & \textbf{0.65} \\
    4 & 0.36 -- 0.54 & 0.45 & 0.62 -- 0.79 & 0.71 & \textbf{0.75} \\
    \bottomrule
    \end{tabular}
    }
\end{table}

\begin{figure}[h]
    \centering
    \includegraphics[width=1\textwidth]{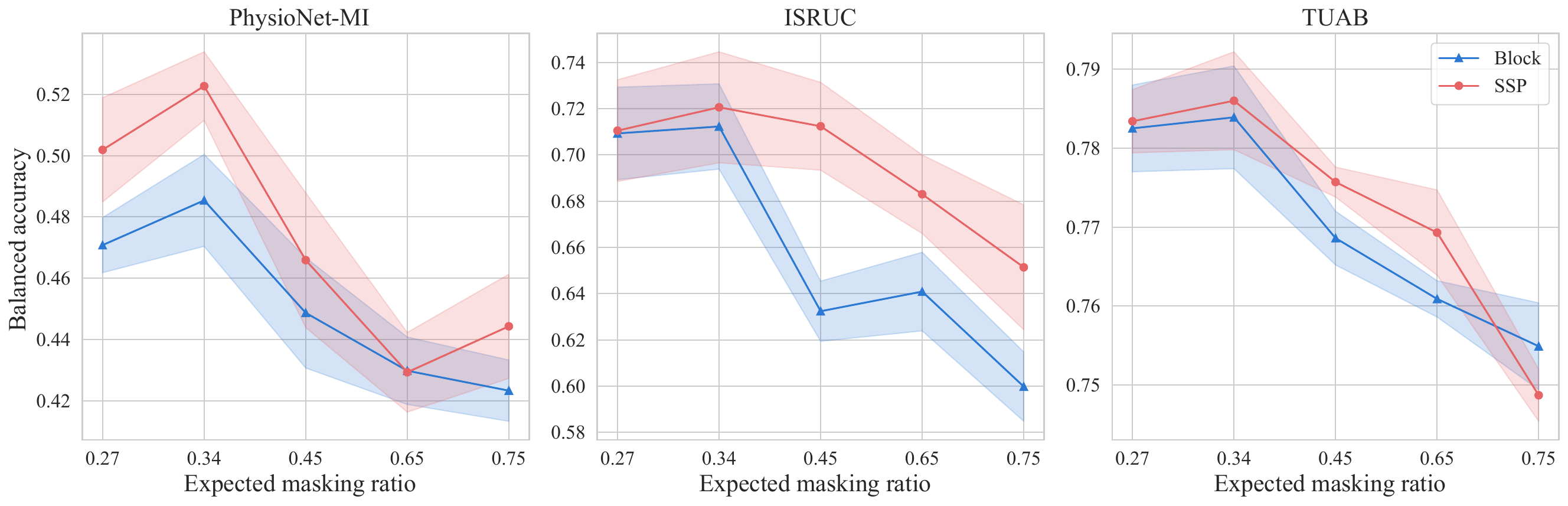}
    \caption{Masking strategy analysis across five difficulty levels on PhysioNet-MI, ISRUC, and TUAB. The blue line represents the performance of block masking, while the red line represents the performance of SSP masking. The shaded areas indicate one standard deviation across five random seeds.}
    \label{fig:masking_results}
\end{figure}

As shown in Figure~\ref{fig:masking_results}, SSP masking generally outperforms block masking, with the best performance on each task achieved under SSP at $\rho \approx$ 0.34.
This suggests that preserving semantic subsequences is beneficial for learning transferable EEG representations.

\subsection{Additional metrics for component ablations}
\label{app:ablation:metrics}

Table~\ref{tab:ablation} reports balanced accuracy only. 
Table~\ref{app:tab:ablation_metrics} adds Cohen's $\kappa$ and weighted F1 for the multi-class tasks and AUC-PR and AUROC for binary TUAB, on the same runs. 
Balanced accuracy is repeated so that the comparison can be read within a single table. 
On each task the two leading variants are the same under every metric.

\begin{table}[ht]
\centering
\caption{Additional metrics for the component ablations of Table~\ref{tab:ablation}. Each cell
reports balanced accuracy followed by two task-appropriate metrics: Cohen's $\kappa$ and weighted
F1 for the multi-class tasks, AUC-PR and AUROC for binary TUAB. The best result per column is in \textbf{bold}; the second-best is \underline{underlined}.}
\label{app:tab:ablation_metrics}
\resizebox{\textwidth}{!}{
\begin{tabular}{lccc}
\toprule
\textbf{Variant} & \textbf{PhysioNet-MI} & \textbf{ISRUC} & \textbf{TUAB} \\
 & (BAcc / $\kappa$ / wF1) & (BAcc / $\kappa$ / wF1) & (BAcc / AUC-PR / AUROC) \\
\midrule
\multicolumn{4}{l}{\textbf{Encoder architecture}} \\
Full attention only & 0.4450 / 0.2600 / 0.4492 & 0.7097 / 0.6314 / 0.7131 & 0.7749 / 0.8538 / 0.8541 \\
Factorized attention only & 0.4833 / 0.3111 / 0.4781 & 0.7153 / 0.6378 / 0.7187 & 0.7865 / 0.8656 / 0.8650 \\
Hybrid + late full & 0.4871 / 0.3161 / 0.4820 & 0.6973 / 0.6194 / 0.7009 & 0.7860 / 0.8651 / 0.8645 \\
\midrule
\multicolumn{4}{l}{\textbf{RSReg}} \\
No RSReg & 0.5182 / 0.3576 / 0.5138 & 0.6994 / 0.6198 / 0.7029 & 0.7759 / 0.8548 / 0.8550 \\
\midrule
\multicolumn{4}{l}{\textbf{Electrode geometry encoding}} \\
Coordinate embedding only & 0.4821 / 0.3095 / 0.4769 & 0.7009 / 0.6215 / 0.7044 & 0.7744 / 0.8533 / 0.8536 \\
Legendre bias only & \underline{0.5193 / 0.3591 / 0.5149} & \textbf{0.7320 / 0.6566 / 0.7352} & \textbf{0.7929 / 0.8722 / 0.8710} \\
Legendre anchor encoding only & 0.5071 / 0.3428 / 0.5025 & 0.7137 / 0.6360 / 0.7171 & 0.7855 / 0.8646 / 0.8640 \\
Learnable pairwise MLP & 0.4463 / 0.2617 / 0.4403 & 0.6679 / 0.5842 / 0.6826 & 0.7671 / 0.8458 / 0.8468 \\
REVE spatial encoding & 0.4921 / 0.3228 / 0.4862 & 0.6910 / 0.6159 / 0.6943 & 0.7607 / 0.8412 / 0.8420 \\
Gaussian (geodesic) & 0.4512 / 0.2683 / 0.4489 & 0.6977 / 0.6189 / 0.7010 & 0.7705 / 0.8501 / 0.8511 \\
\midrule
SPERA-Small & \textbf{0.5227 / 0.3636 / 0.5184} & \underline{0.7206 / 0.6438 / 0.7239} & \underline{0.7869 / 0.8660 / 0.8653} \\
\bottomrule
\end{tabular}}
\end{table}

\section{Analysis of the Legendre spherical prior}
\label{app:prior}

\subsection{Geometry encoding ablations}
\label{app:prior:ablation}
Building on the electrode geometry ablation in Section~\ref{sec:ablation}, we evaluate a broader set of alternative geometry encodings.
Everything outside the geometry encoding is held fixed.
\emph{Coordinate embedding only} removes the pairwise prior and retains the linear embedding of 3D electrode coordinates. 
\emph{Anisotropic SH} replaces the rotation-invariant Legendre expansion with a general spherical-harmonic (SH) prior that permits direction-dependent interactions and contains the isotropic Legendre family as a special case.
\emph{Gaussian (geodesic)} replaces the expansion with a radial basis function (RBF) kernel of geodesic electrode distance, with learnable width and amplitude. 
\emph{REVE 4D encoding} ports the coupled spatio-temporal encoding of REVE~\cite{ouahidi2025reve} as published, which also replaces our temporal RoPE. 
Since that conflates spatial and temporal handling, we derive \emph{REVE spatial encoding}, which imports only the 3D coordinate component and leaves RoPE in place, isolating the geometry comparison.
 
The results are summarized in Table~\ref{app:tab:geometry}.
The Legendre prior is the strongest configuration on all three tasks, and REVE's encoding provides no transfer advantage in our backbone. 
Replacing REVE's joint 4D encoding with its spatial component while retaining RoPE improves performance on all three tasks, with the largest increase on ISRUC (0.593 to 0.691).
At matched temporal handling, REVE's Fourier encoding is comparable to our linear coordinate embedding, and the improvement comes from adding the prior on top.
The geodesic Gaussian fixes the kernel to a monotone decay in distance, whereas our expansion learns its profile end-to-end; the learned kernel is not monotone (Appendix~\ref{app:prior:coeff}). 
The anisotropic SH prior also underperforms the isotropic Legendre prior (0.658 vs. 0.677 average), providing no empirical benefit from the added directional flexibility under the present setting.
 
\begin{table}[t]
\centering
\caption{Geometry encoding ablations. All variants use the SPERA-Small backbone under the ablation
protocol of Section~\ref{sec:ablation}. Results are mean balanced accuracy $\pm$ standard
deviation over five random seeds; the best result in each column is in bold.}
\label{app:tab:geometry}
\resizebox{\textwidth}{!}{
\begin{tabular}{lcccc}
\toprule
\textbf{Geometry encoding} & \textbf{PhysioNet-MI} & \textbf{ISRUC} & \textbf{TUAB} & \textbf{Average} \\
\midrule
Anisotropic SH               & $0.501 \pm 0.013$ & $0.706 \pm 0.017$ & $0.767 \pm 0.008$ & $0.658 \pm 0.013$ \\
Coordinate embedding only    & $0.482 \pm 0.015$ & $0.701 \pm 0.027$ & $0.774 \pm 0.006$ & $0.653 \pm 0.010$ \\
REVE spatial encoding        & $0.492 \pm 0.017$ & $0.691 \pm 0.021$ & $0.761 \pm 0.009$ & $0.648 \pm 0.009$ \\
Gaussian (geodesic)          & $0.451 \pm 0.016$ & $0.698 \pm 0.023$ & $0.771 \pm 0.009$ & $0.640 \pm 0.010$ \\
REVE 4D encoding             & $0.445 \pm 0.015$ & $0.593 \pm 0.028$ & $0.712 \pm 0.010$ & $0.583 \pm 0.017$ \\\midrule
Legendre prior (SPERA-Small) & \textbf{0.523 $\pm$ 0.011} & \textbf{0.721 $\pm$ 0.024} & \textbf{0.787 $\pm$ 0.006} & \textbf{0.677 $\pm$ 0.011} \\
\bottomrule
\end{tabular}}
\end{table}

\subsection{Learned coefficients and angular bias profile}
\label{app:prior:coeff}

To examine what spatial structure is learned by the pairwise Legendre bias, we analyze the coefficients of the pretrained SPERA-Base model before downstream fine-tuning. The pairwise bias of Eq.~(2) can be written as a rotation-invariant angular bias function

\begin{equation}
    K(\theta) = \sum_{\ell=0}^{L} w_\ell P_\ell(\cos\theta),
\end{equation}
where $\theta$ denotes the angular separation between two electrodes. Because $P_0(\cos\theta)=1$, the degree-0 term contributes only an angle-independent offset. We therefore exclude it when analyzing the relative shape of the learned spatial kernel.

As shown in Table~\ref{app:tab:legendre_coeff}, the learned coefficients are concentrated at low Legendre degrees. Degrees 1--3 account for 93.1\% of the summed squared coefficients over degrees 1--8, indicating that the learned bias is dominated by broad, smoothly varying angular structure. Evaluating $K(\theta)$ over angular separations from $0^\circ$ to $180^\circ$ using the learned coefficients shows that the bias reaches its maximum at approximately $20.0^\circ$ and its minimum at $180^\circ$, assigning relatively higher bias to nearby electrode pairs than to widely separated or antipodal pairs. The nonzero location of the maximum also shows that this locality-oriented profile is not a simple monotonic decay with angular distance.

\begin{table}[h]
\centering
\caption{Learned pairwise Legendre coefficients of the pretrained SPERA-Base model. Degree 0 is omitted because it contributes only an angle-independent offset.}
\label{app:tab:legendre_coeff}
\resizebox{\textwidth}{!}{%
\begin{tabular}{lrrrrrrrr}
\toprule
\textbf{Legendre degree} & 1 & 2 & 3 & 4 & 5 & 6 & 7 & 8 \\
\midrule
Coefficient & 0.1332 & 0.0831 & 0.0415 & 0.0074 & $-$0.0003 & $-$0.0205 & $-$0.0089 & $-$0.0376 \\
\bottomrule
\end{tabular}}
\end{table}

\subsection{Decomposition of the anchor-based attention}
\label{app:prior:decomp}

Unlike the pairwise Legendre bias of Eq.~(2), the anchor-based formulation injects the geometry feature $\boldsymbol{\phi}_c$ into the queries and keys. Expanding Eq.~(4) therefore yields four score components:

\begin{equation}
\begin{split}
s_{ij} &= \frac{1}{\sqrt{d}}
\big( \mathbf{q}_i + \Gamma_Q\boldsymbol{\phi}_{c_i} \big)^{\!\top}
\big( \mathbf{k}_j + \Gamma_K\boldsymbol{\phi}_{c_j} \big) \\
&= \frac{1}{\sqrt{d}} \Big(
\underbrace{\mathbf{q}_i^\top \mathbf{k}_j}_{\text{content--content}}
+ \underbrace{\mathbf{q}_i^\top \Gamma_K\boldsymbol{\phi}_{c_j}
+ (\Gamma_Q\boldsymbol{\phi}_{c_i})^\top \mathbf{k}_j}_{\text{content--geometry}}
+ \underbrace{(\Gamma_Q\boldsymbol{\phi}_{c_i})^\top \Gamma_K\boldsymbol{\phi}_{c_j}}_{\text{geometry--geometry}}
\Big) .
\end{split}
\end{equation}

The geometry--geometry term captures pairwise interaction between the Legendre anchor encodings, while the two cross terms couple token content with electrode geometry. These cross terms have no counterpart in Eq.~(2), where geometry enters only as an additive pairwise score bias.

We measure the contribution of each term to the pre-softmax logit using its root-mean-square magnitude over randomly sampled token pairs and attention heads in each full-attention block. The two content--geometry cross terms together exceed the geometry--geometry term in every full-attention block of both SPERA-Base and SPERA-Large, by factors of 1.3 to 13. Content--content remains the largest single term except in the earliest full-attention block, where the two cross terms combined exceed it in both models.

\subsection{Pairwise-bias and anchor-based implementations}
\label{app:prior:impl}

The two implementations differ in how geometry enters attention and therefore in computational cost. 
In the pairwise Legendre bias form, the Legendre values $P_\ell(\mathbf{e}_i^\top\mathbf{e}_j)$ are computed once per forward pass and reused, so the overhead does not arise from evaluating the polynomials.
Instead, adding the pairwise Legendre bias to the attention scores requires an $N\times N$ bias matrix and prevents the fused attention path used in our implementation. 
We test whether this overhead can be reduced using PyTorch's FlexAttention, whose \texttt{score\_mod} applies the pairwise Legendre bias inside the fused kernel without materializing the full score matrix.

Table~\ref{app:tab:throughput} compares three whole-model configurations of SPERA-Small.
Throughput increases from 52,263 to 66,326 tok/s with FlexAttention, but remains about 29\% below that of the anchor-based design.
Since the anchor form costs only 0.005 average balanced accuracy relative to the pairwise bias (Table~\ref{tab:ablation}), we retain it in the full-attention blocks.
 
\begin{table}[ht]
\centering
\caption{Training throughput for three whole-model configurations of SPERA-Small, measured under the setting of Table~\ref{tab:efficiency_profiling}. Only the treatment of the prior in full-attention blocks differs; factorized layers use the pairwise Legendre bias throughout.}
\label{app:tab:throughput}
\begin{tabular}{llc}
\toprule
\textbf{Prior in full-attention blocks} & \textbf{Throughput (tok/s)} \\
\midrule
Pairwise Legendre bias                      & 52,263 \\
Pairwise Legendre bias with FlexAttention   & 66,326 \\
Legendre anchor encoding (SPERA)      & 92,866 \\
\bottomrule
\end{tabular}
\end{table}

\subsection{Extension to intracranial recordings}
\label{app:prior:ieeg}
 
For scalp EEG, electrodes are distributed over the scalp surface, and we therefore represent their locations using unit-normalized directions. 
For iEEG and sEEG, where electrodes occupy the brain volume, radial position also becomes informative.
The construction can be extended to the solid ball $\mathbb{B}^3$ while retaining the same rotation-invariant angular structure. 
On the solid ball, separable solutions of the Laplacian eigenproblem consist of a radial part given by spherical Bessel functions $j_\ell$ and an angular spherical-harmonic part. 
Using the corresponding rotation-invariant angular form yields
\begin{equation}
k(\mathbf{x}_i, \mathbf{x}_j) = \sum_{\ell, n} w_{\ell n}\, j_\ell(\kappa_n r_i)\, j_\ell(\kappa_n r_j)\,
P_\ell(\mathbf{e}_i^\top \mathbf{e}_j) ,
\end{equation}
where $r_i$ and $\mathbf{e}_i$ are the radius and unit direction of electrode $i$, $n$ indexes radial modes, $\kappa_n$ denotes the radial wavenumber, and $w_{\ell n}$ are learnable coefficients.
The angular Legendre structure is retained and radius enters through a learnable radial basis; the surface prior used in this paper is the special case of a shared radius.
 
Two practical obstacles remain: intracranial electrode positions require coregistration to a common anatomical space, such as the Montreal Neurological Institute (MNI) template, before a shared radial basis is meaningful, and the high-gamma content that carries much of the information in intracranial recordings lies outside our 0.5--75~Hz preprocessing pipeline, so the tokenizer would need revision.
We leave this extension to future work.

\subsection{Spatial attribution analysis}
\label{app:spatial:attribution}
We examine whether the model's electrode-level attributions align with known EEG neurophysiological patterns.
We use integrated gradients (IG)~\cite{mukund2017axiomatic} on the fine-tuned SPERA-Base model, quantifying each electrode's contribution to the classification.
We focus on motor imagery (MI), for which the relevant cortical regions and class-dependent spatial organization are well established~\cite{pfurtscheller2001motor}.

IG is computed against a zero baseline on the test split.
Per-class IG maps are strongly correlated (mean pairwise $r$ = 0.68 on PhysioNet-MI, and 0.66 on BCIC-IV-2a), being dominated by a shared channel-saliency profile that concentrates on the sensorimotor electrodes: mean absolute IG is 3.5--4.5 at C3, C4, CP3, and CP4 on PhysioNet-MI against 1.7--2.4 elsewhere, and 25--41 at CP3, CP4, FC3, and FC4 on BCIC-IV-2a against 15.7--17.5 in frontal and parietal regions.
This shared profile is consistent with reliance on signals recorded over sensorimotor scalp regions.
To highlight class-relative spatial differences, we apply a two-step normalization: each class map is divided by its mean over channels, and each channel is then divided by its mean over classes.
In these normalized maps, 1.0 corresponds to the across-class mean for each channel.
We summarize the class-relative maps over the sensorimotor strip (all FC*, C*, and CP* electrodes), split into left, right, and midline groups, using a lateralization index LI = (right $-$ left) / (right + left) and a mediality index MedIdx = midline / mean (left, right). 

\begin{table}[t]
\centering
\caption{Class-relative attribution over the sensorimotor strip for BCIC-IV-2a and PhysioNet-MI. Each value is the class-relative IG after the two-step normalization.
LI $=$ (SM-right $-$ SM-left) / (SM-right $+$ SM-left);
MedIdx $=$ SM-mid $/$ mean (SM-left, SM-right).}
\label{tab:class_rel_ig}
\begin{tabular}{llccccc}
\toprule
\textbf{Dataset} & \textbf{Class} & \textbf{SM-left} & \textbf{SM-right}
 & \textbf{SM-mid} & \textbf{LI} & \textbf{MedIdx} \\
\midrule
\multirow{4}{*}{BCIC-IV-2a}
 & Left hand  & 0.953 & 1.095 & 0.959 & $+$0.069 & 0.937 \\
 & Right hand & 1.081 & 0.991 & 0.872 & $-$0.044 & 0.841 \\
 & Feet       & 0.971 & 0.936 & 1.133 & $-$0.018 & 1.187 \\
 & Tongue     & 0.994 & 0.978 & 1.036 & $-$0.008 & 1.051 \\
\midrule
\multirow{4}{*}{PhysioNet-MI}
 & Left fist  & 0.942 & 1.123 & 0.980 & $+$0.088 & 0.949 \\
 & Right fist & 1.015 & 0.992 & 0.803 & $-$0.012 & 0.800 \\
 & Both fists & 0.988 & 0.915 & 1.088 & $-$0.038 & 1.144 \\
 & Both feet  & 1.056 & 0.970 & 1.129 & $-$0.042 & 1.115 \\
\bottomrule
\end{tabular}
\end{table}

\begin{figure}[t]
    \centering
    \includegraphics[width=1.0\textwidth]{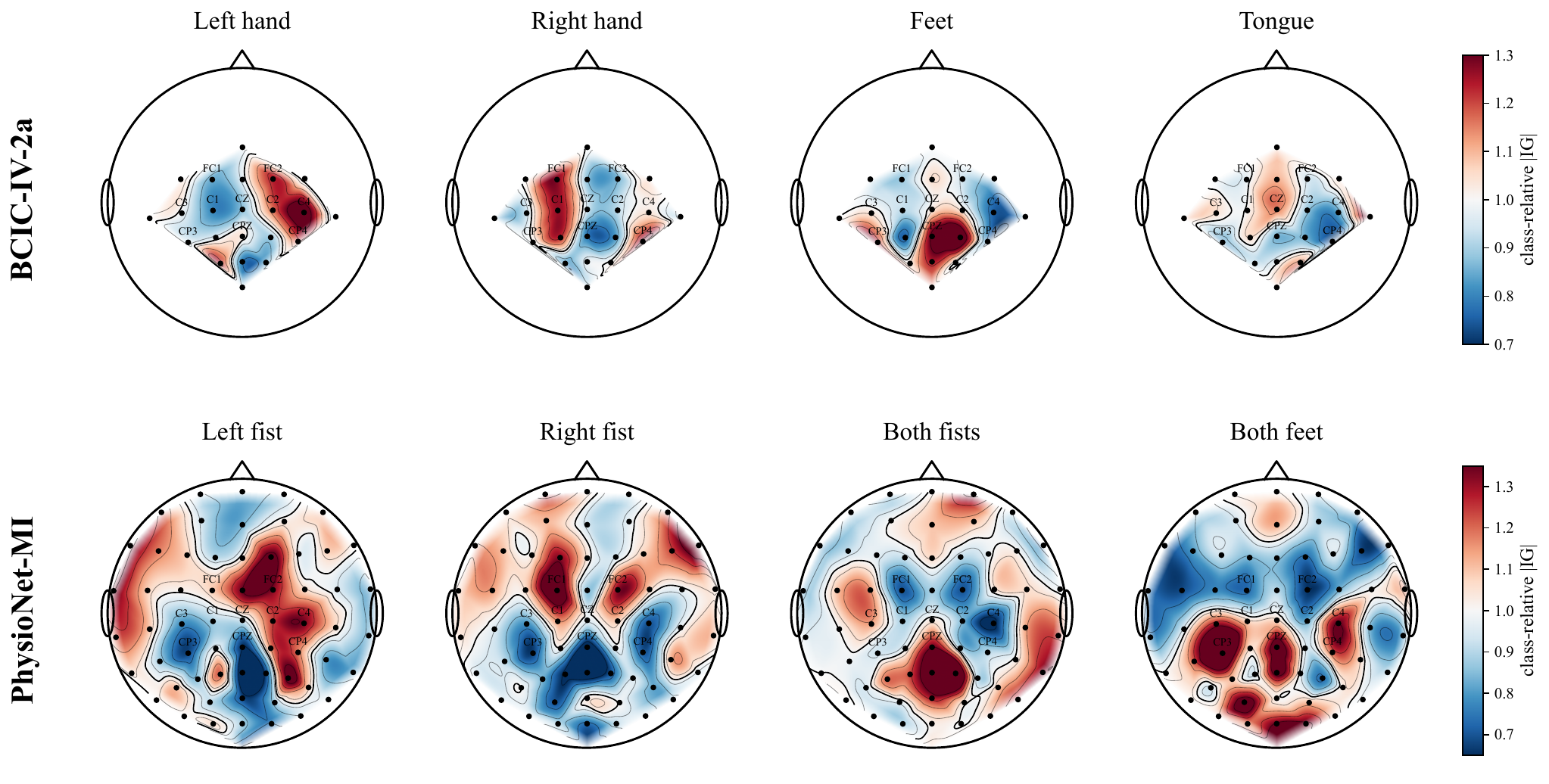}
    \caption{Class-relative attribution topographic maps on BCIC-IV-2a and PhysioNet-MI. Maps are computed separately for each class after two-step class-relative normalization.}
    \label{fig:rel_IG_topo}
\end{figure}

On BCIC-IV-2a, the two unimanual classes follow the expected pattern (Table~\ref{tab:class_rel_ig}): LI is positive for left-hand and negative for right-hand imagery, and Figure~\ref{fig:rel_IG_topo} shows the corresponding increase over the sensorimotor region contralateral to the imagined hand, consistent with the established lateralization of hand motor imagery~\cite{pfurtscheller2006mu, neuper2006erd}. 
The feet class has the highest mediality and a correspondingly central pattern around the midline, reflecting the medial cortical representation of the lower limbs~\cite{pfurtscheller2006mu, penfield1937somatic}. 
The tongue class has the smallest absolute LI of the four and is broadly distributed across the sensorimotor strip without the unilateral concentration seen for the hand classes, in line with reports of distributed parietofrontal involvement during tongue motor imagery~\cite{giannopulu2021neural}.

PhysioNet-MI shows the same organization, though more weakly for the right-fist class: attribution shifts toward the right hemisphere for the left fist and only marginally toward the left hemisphere for the right fist. 
Both bilateral classes have MedIdx above 1 while both unimanual classes fall below it, and the topographic maps place their attribution around central and midline sensorimotor sites. 

\section{Few-shot experiments}
\label{app:fewshot}
To evaluate the data efficiency of SPERA, we conduct few-shot LP experiments on BCIC-IV-2a, PhysioNet-MI, and ISRUC (motor imagery and sleep staging).
For each dataset, the encoder is frozen and only a linear classifier is trained using a limited subset of the original training split. 
We evaluate six training data ratios: 1\%, 2\%, 5\%, 10\%, 25\%, and 50\%. 
The subsets are sampled in a class-balanced manner based on the smallest class size in the training set. 
All experiments are repeated over five random seeds, and we report the mean balanced accuracy with standard deviation.

\begin{figure}[ht]
    \centering
    \includegraphics[width=1.0\textwidth]{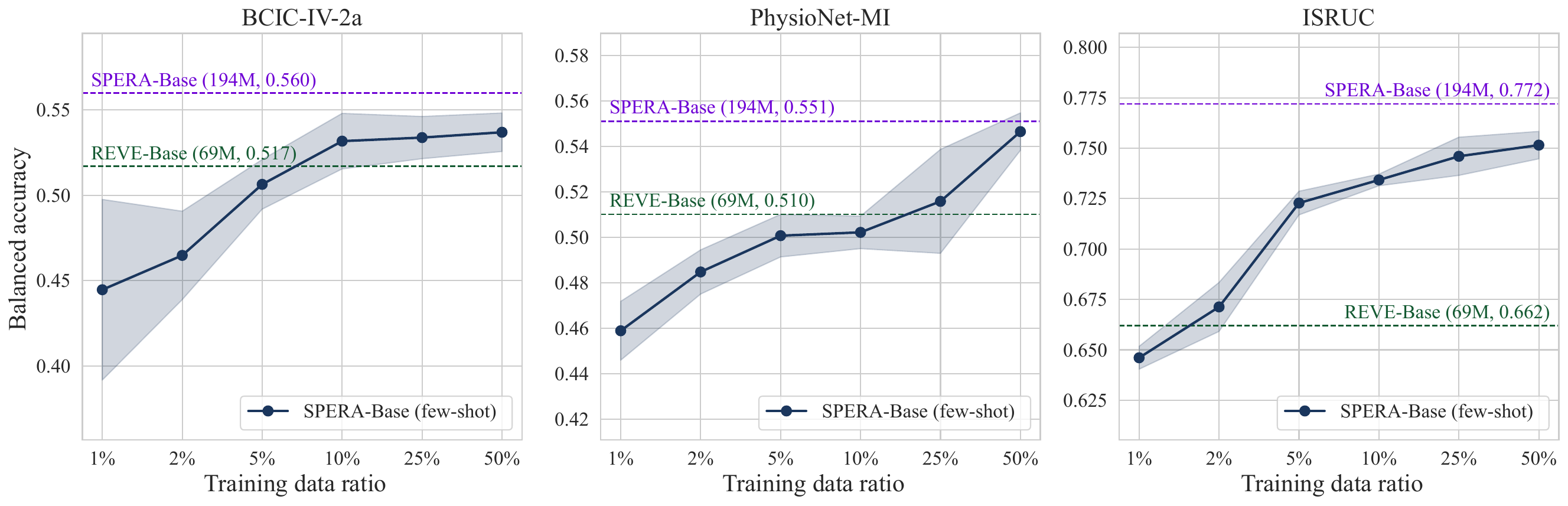}
    \caption{
    Few-shot LP results of SPERA-Base on BCIC-IV-2a, PhysioNet-MI, and ISRUC. 
    The blue curve shows the mean balanced accuracy of SPERA-Base under limited labeled training data, with the shaded area indicating one standard deviation across five random seeds. 
    Dashed horizontal lines indicate the full-data LP results of SPERA-Base and REVE-Base.
    }
    \label{fig:fewshot_results}
\end{figure}

Across all three datasets, performance improves steadily as the training ratio increases (Figure~\ref{fig:fewshot_results}). 
On BCIC-IV-2a, SPERA-Base surpasses the REVE-Base reference using 10\% of the training data. 
PhysioNet-MI shows a more gradual trend, with the 50\% setting closing the gap to the full-data SPERA-Base reference.
On ISRUC, SPERA-Base exceeds the REVE-Base reference with only 2\% of the training data and approaches the full-data SPERA-Base reference at higher ratios.
These results indicate that SPERA provides data-efficient EEG representations.

\section{Noise robustness analysis}
\label{app:noise}

We evaluate noise robustness under LP using SPERA-Small, CBraMod, and REVE-Base. To isolate the effect of the pretraining objective, we additionally include the matched reconstruction variant from Appendix~\ref{app:ablation:objective}, which differs only in the prediction target. Since CBraMod and REVE do not report noise robustness results, we re-evaluate their publicly released checkpoints together with the noise-free condition under the same protocol. For each checkpoint, the encoder is frozen and the linear classifier and feature-normalization statistics are fitted only on unperturbed training features and then kept fixed across test conditions. We consider additive Gaussian noise and injected ocular artifacts separately.

\subsection{Gaussian noise}
Gaussian noise robustness is evaluated on TUAB, ISRUC, and PhysioNet-MI. 
Additive Gaussian white noise is applied only to the test inputs under five conditions: noise-free, 10 dB, 5 dB, 0 dB, and $-$5 dB.
Noise variance is determined independently for each channel from its mean squared signal amplitude to match the target signal-to-noise ratio (SNR).

\begin{figure}[h]
    \centering
    \includegraphics[width=1\textwidth]{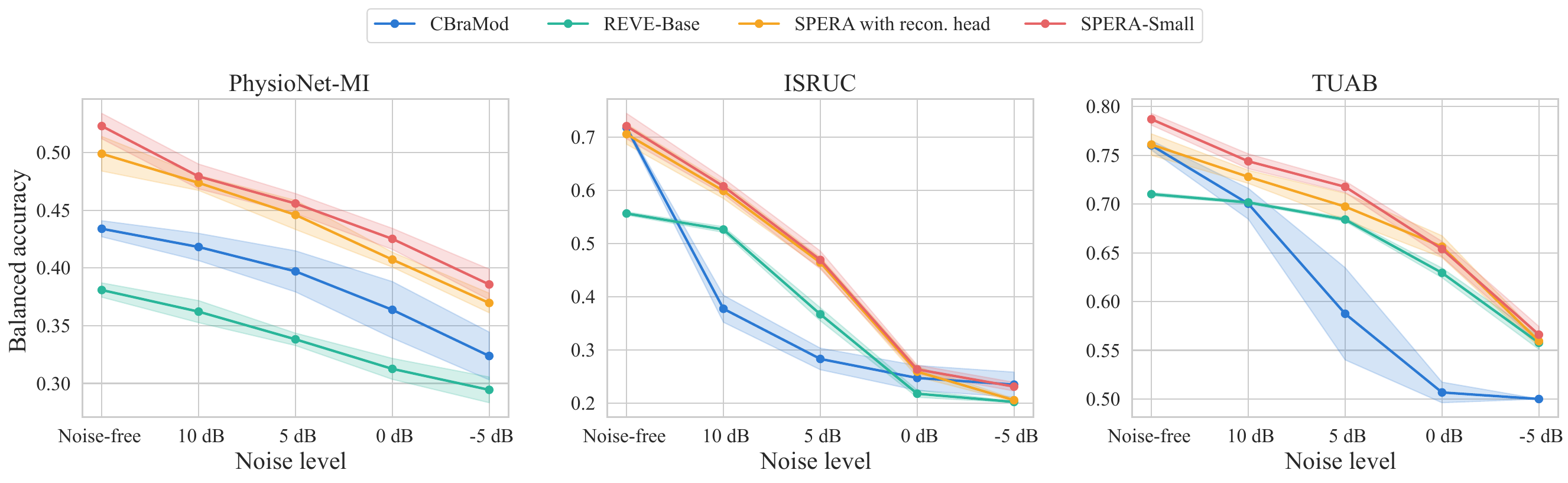}
    \caption{Gaussian noise robustness on PhysioNet-MI, ISRUC, and TUAB under LP. Frozen encoders with linear classifiers trained on unperturbed data are evaluated on test inputs with additive Gaussian noise at 10, 5, 0, and $-5$ dB. Shaded regions denote one standard deviation across five random seeds.}
    \label{fig:noise_gaussian}
\end{figure}

As shown in Figure~\ref{fig:noise_gaussian}, SPERA-Small generally maintains higher mean balanced accuracy than the comparison models across the evaluated noise conditions. The matched JEPA and reconstruction variants show broadly similar degradation patterns, while JEPA maintains higher mean accuracy in nearly all evaluated conditions. Thus, the accuracy advantage of latent prediction generally persists under additive Gaussian corruption.

\subsection{Physiological ocular artifacts}

We further evaluate robustness to ocular artifacts on PhysioNet-MI and TUAB using electrooculographic (EOG)  waveforms from EEGdenoiseNet~\cite{Zhang_2021}; ISRUC is excluded because the ocular artifact injection protocol is not appropriate for sleep EEG. Ocular artifacts are projected onto each montage using the empirical propagation factors reported by Lins et al.~\cite{lins1993ocular}. For TUAB, bipolar-channel factors are obtained by subtracting those of the constituent electrodes; for PhysioNet-MI, factors are estimated by three-dimensional thin-plate spline interpolation with a linear term, enforcing symmetry for the vertical field and antisymmetry for the horizontal field. For each trial, independently sampled vertical and horizontal EOG waveforms are normalized, projected using these spatial fields, and combined using the reported horizontal-to-vertical saccade amplitude ratio. A shared intermittent envelope leaves at least 30\% of each trial without added artifacts, after which the combined artifact is scaled to the target SNR based on trial-level mean EEG and artifact power across channels and time.

\begin{figure}[b]
    \centering
    \includegraphics[width=1\textwidth]{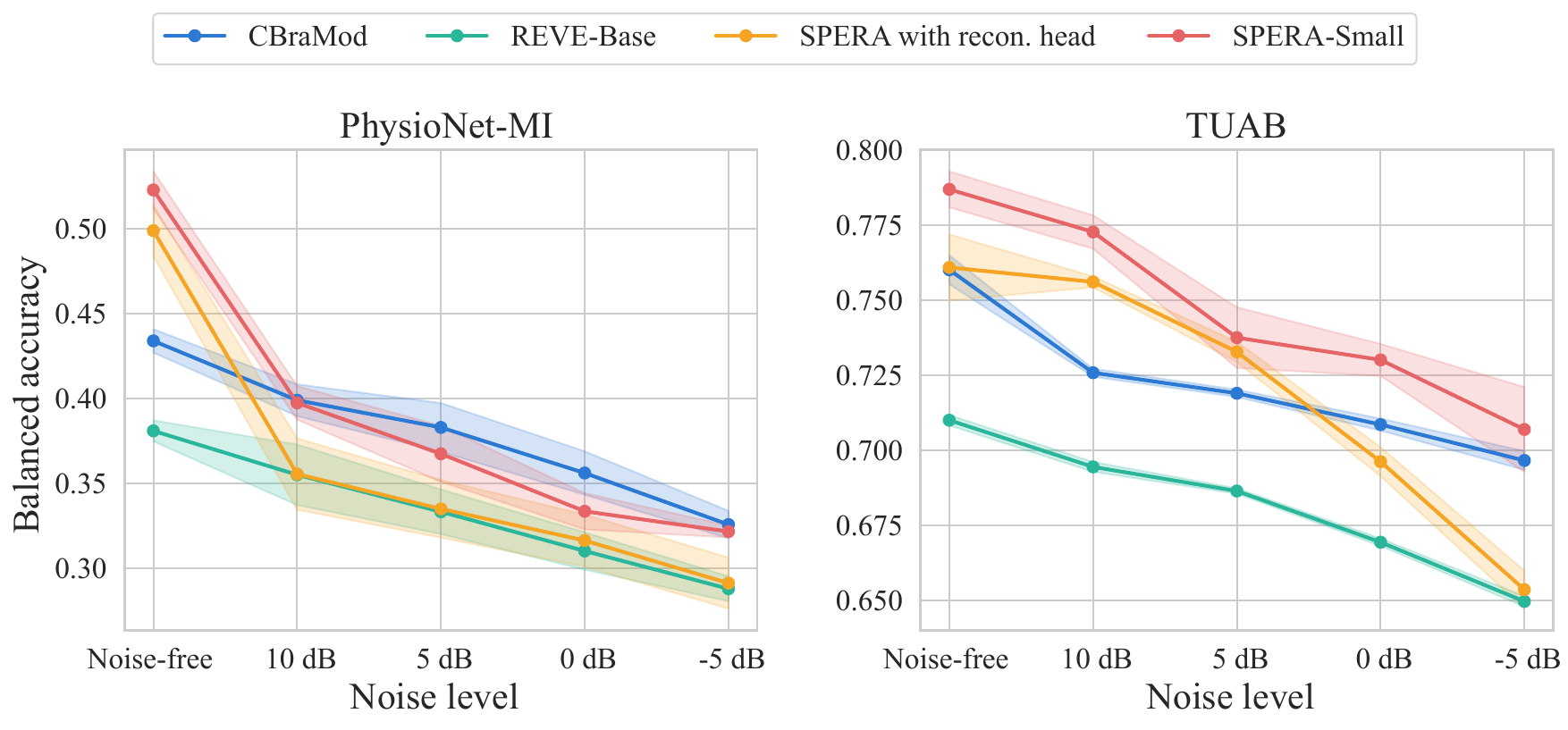}
    \caption{Ocular-artifact robustness on PhysioNet-MI and TUAB under LP. Frozen encoders with linear classifiers trained on unperturbed data are evaluated on test inputs with injected EEGdenoiseNet EOG artifacts at 10, 5, 0, and $-5$ dB. Shaded regions denote one standard deviation across five random seeds.}
    \label{fig:noise_eog}
\end{figure}

Figure~\ref{fig:noise_eog} shows task-dependent robustness patterns under injected ocular artifacts. On TUAB, SPERA-Small achieves the highest mean balanced accuracy across all conditions, whereas CBraMod performs best under the injected-artifact conditions on PhysioNet-MI. In the matched objective comparison, JEPA remains above reconstruction under every condition on both datasets. From noise-free to $-$5 dB, the decrease is smaller for JEPA on PhysioNet-MI (0.201 vs.\ 0.208) and TUAB (0.080 vs.\ 0.107), with the larger difference observed on TUAB.

\section{Representation analysis}
\label{app:representation_analysis}
\subsection{UMAP visualization of learned representations}
\label{app:vis}

\begin{figure}[ht]
    \centering
    \includegraphics[width=1\textwidth]{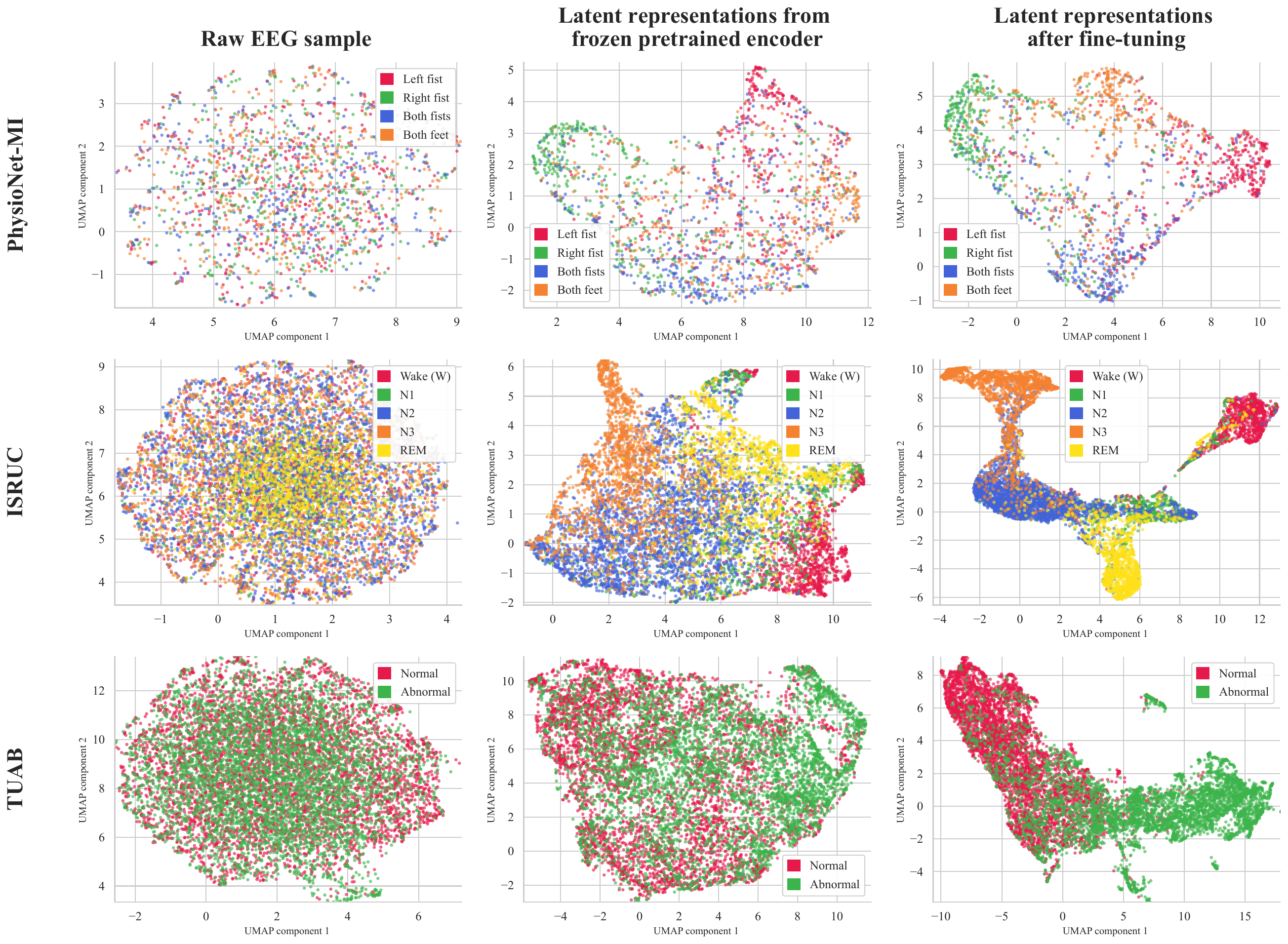}
    \caption{UMAP visualization of representations from the SPERA-Base model on three downstream tasks. Columns correspond to raw EEG, frozen pretrained features, and fine-tuned features. Rows correspond to PhysioNet-MI, ISRUC, and TUAB.}
    \label{fig:representation}
\end{figure}

To qualitatively examine the structure of learned representations, we apply uniform manifold approximation and projection (UMAP)~\cite{mcinnes2018umap} to PhysioNet-MI, ISRUC, and TUAB. 
For each dataset, we visualize three representation levels: (i) raw EEG samples, (ii) features from the frozen pretrained encoder, and (iii) features after task-specific fine-tuning. 
The results are shown in Figure~\ref{fig:representation}.

The first column shows that raw EEG samples exhibit no clear class-wise structure across all three datasets. 
The second column reveals that the frozen pretrained features already capture task-related structure to varying degrees. 
On ISRUC, sleep stages exhibit clearly distinguishable clusters even without task-specific supervision, with Wake, N3, and REM occupying separate areas of the embedding space. 
TUAB shows a weaker but visible separation between normal and abnormal recordings, while PhysioNet-MI exhibits more limited class-level structure at this stage. 
The third column shows that, after fine-tuning, all three datasets form clearer and more separable clusters.

\subsection{Linear accessibility and encoder adaptation}

In terms of average balanced accuracy across the nine tasks, SPERA-Base leads REVE-Base by 0.051 under LP and by 0.035 under LPFT (Tables~\ref{tab:main} and~\ref{tab:lp}). 
At the aggregate level, SPERA-Base remains ahead under both protocols, although several task-level differences vary with model scale and adaptation. 
Under LP, REVE-Base leads SPERA-Base on MAT, while REVE-Large leads SPERA-Large on BCIC-IV-2a, PhysioNet-MI, and Mumtaz. 
BCIC-IV-2a and PhysioNet-MI are motor-imagery tasks that involve frequency-specific sensorimotor activity~\cite{pfurtscheller1999event}, while MAT has been used to study spectral changes during mental arithmetic~\cite{zyma2019electroencephalograms}, motivating us to examine the linear accessibility of band-power information.

We regress channel-wise relative power in five frequency bands (delta to gamma) from frozen encoder representations using a linear readout on per-channel token features. 
Table~\ref{app:tab:bandpower_decodability} reports held-out macro Pearson's $r$, averaged over channel--band targets. 
Because SPERA and REVE differ in architecture, tokenization, corpus, and pretraining objective, we additionally evaluate the matched reconstruction variant of SPERA-Small from Appendix~\ref{app:ablation:objective}.

\begin{table}[h]
\centering
\caption{Band-power decodability from frozen representations. Values are held-out macro Pearson's $r$, averaged over channel--band targets.}
\label{app:tab:bandpower_decodability}
\begin{tabular}{lcccc}
\toprule
\textbf{Method} & \textbf{BCIC-IV-2a} & \textbf{PhysioNet-MI} & \textbf{MAT} & \textbf{Average} \\
\midrule
SPERA-Small & 0.395 & 0.482 & 0.305 & 0.394 \\
Reconstruction head & 0.592 & 0.640 & 0.483 & 0.572 \\
REVE-Base & 0.735 & 0.680 & 0.589 & 0.668 \\
\bottomrule
\end{tabular}
\end{table}

Across all three tasks, band power is more linearly decodable from the matched reconstruction variant than from SPERA-Small, with REVE-Base showing the highest values.
On PhysioNet-MI, however, the same objective swap increases band-power decodability from 0.482 to 0.640 while lowering LP accuracy from 0.523 to 0.499 (Appendix~\ref{app:ablation:objective}), indicating that greater band-power accessibility alone does not explain the downstream accuracy difference.

To further investigate the LP--LPFT divergence, we test how much of the performance gap can be recovered by limited encoder adaptation. Unfreezing only the last of SPERA-Base's 16 encoder blocks after LP recovers 49\%, 71\%, and 61\% of the LP--LPFT gain on BCIC-IV-2a (0.560 to 0.619), PhysioNet-MI (0.551 to 0.637), and MAT (0.672 to 0.736), respectively. 
These results suggest that a substantial portion of the LP--LPFT gap can be recovered through limited encoder adaptation.

%%%%%%%%%%%%%%%%%%%%%%%%%%%%%%%%%%%%%%%%%%%%%%%%%%%%%%%%%%%%

\newpage
\section*{NeurIPS Paper Checklist}

\begin{enumerate}

\item {\bf Claims}
    \item[] Question: Do the main claims made in the abstract and introduction accurately reflect the paper's contributions and scope?
    \item[] Answer: \answerYes{} % Replace by \answerYes{}, \answerNo{}, or \answerNA{}.
    \item[] Justification: The abstract and Section~\ref{sec:intro} describe SPERA as an EEG representation-learning framework based on latent-space prediction, with a Legendre-polynomial spherical prior for electrode geometry, a hybrid attention backbone, and a relational spectral regularizer. The empirical claims regarding downstream performance, scaling and parameter efficiency, and robustness are supported by Sections~\ref{sec:downstream_results}--\ref{sec:ablation} and Appendices~\ref{app:down}--\ref{app:representation_analysis}.

\item {\bf Limitations}
    \item[] Question: Does the paper discuss the limitations of the work performed by the authors?
    \item[] Answer: \answerYes{} % Replace by \answerYes{}, \answerNo{}, or \answerNA{}.
    \item[] Justification: Section~\ref{sec:limitations} discusses limitations regarding (i) scaling beyond 396M parameters, (ii) the lack of systematic disentanglement of corpus size, model capacity, and architectural priors in the observed scaling behavior, (iii) the surface-electrode assumption of the spherical prior, (iv) evaluation restricted to classification over fixed-duration EEG segments of 3--30 seconds, and (v) incomplete demographic metadata for comprehensive fairness analysis.

\item {\bf Theory assumptions and proofs}
    \item[] Question: For each theoretical result, does the paper provide the full set of assumptions and a complete (and correct) proof?
    \item[] Answer: \answerNA{} % Replace by \answerYes{}, \answerNo{}, or \answerNA{}.
    \item[] Justification: The paper presents an empirical contribution and does not include theoretical results such as theorems, propositions, or lemmas requiring formal proof.

    \item {\bf Experimental result reproducibility}
    \item[] Question: Does the paper fully disclose all the information needed to reproduce the main experimental results of the paper to the extent that it affects the main claims and/or conclusions of the paper (regardless of whether the code and data are provided or not)?
    \item[] Answer: \answerYes{} % Replace by \answerYes{}, \answerNo{}, or \answerNA{}.
    \item[] Justification: Section~\ref{sec:methods} specifies the model architecture, training objectives, masking strategy, and augmentation, while Sections~\ref{sec:pretraining}--\ref{sec:eval_protocol} describe the pretraining corpus, preprocessing, and downstream evaluation. Appendices~\ref{app:computing_resources}--\ref{app:down} provide computing resources, corpus information, hyperparameters, and dataset-specific evaluation details, and Appendices~\ref{app:ablations}--\ref{app:representation_analysis} document the protocols for the additional ablations and analyses.

\item {\bf Open access to data and code}
    \item[] Question: Does the paper provide open access to the data and code, with sufficient instructions to faithfully reproduce the main experimental results, as described in supplemental material?
    \item[] Answer: \answerYes{} % Replace by \answerYes{}, \answerNo{}, or \answerNA{}.
    \item[] Justification: Our codebase is provided as a link in the paper (\url{https://github.com/sees4321/SPERA_EEG}), including training, evaluation, and preprocessing scripts together with a README detailing the environment setup and exact commands to reproduce the main results. All pretraining and downstream datasets are publicly available, with sources and licenses listed in Appendix~\ref{app:ptds}. 

\item {\bf Experimental setting/details}
    \item[] Question: Does the paper specify all the training and test details (e.g., data splits, hyperparameters, how they were chosen, type of optimizer) necessary to understand the results?
    \item[] Answer: \answerYes{} % Replace by \answerYes{}, \answerNo{}, or \answerNA{}.
    \item[] Justification: Section~\ref{sec:pretraining} specifies the optimizer and pretraining setup, and Section~\ref{sec:eval_protocol} describes SPERA's LP and LPFT protocols, validation-based hyperparameter selection, and baseline evaluation settings. Appendix~\ref{app:impl} provides the full hyperparameters across all model sizes, while Appendix~\ref{app:down} documents the dataset-specific preprocessing and train/validation/test splits, including implementation-specific details.

\item {\bf Experiment statistical significance}
    \item[] Question: Does the paper report error bars suitably and correctly defined or other appropriate information about the statistical significance of the experiments?
    \item[] Answer: \answerYes{} % Replace by \answerYes{}, \answerNo{}, or \answerNA{}.
    \item[] Justification: SPERA experiments are repeated over five random seeds and reported as mean $\pm$ one standard deviation; for the aggregate results, variability is computed across the corresponding seed-wise averages. Appendix~\ref{app:significance} additionally reports unpaired Welch's $t$-tests with 95\% confidence intervals and Holm correction across the nine tasks separately for each SPERA model, while figure captions specify shaded one-standard-deviation bands where applicable.

\item {\bf Experiments compute resources}
    \item[] Question: For each experiment, does the paper provide sufficient information on the computer resources (type of compute workers, memory, time of execution) needed to reproduce the experiments?
    \item[] Answer: \answerYes{} % Replace by \answerYes{}, \answerNo{}, or \answerNA{}.
    \item[] Justification: Appendix~\ref{app:computing_resources} details the computing environment, including the hardware (HP Z8 workstation with two NVIDIA RTX 5000 Ada GPUs), software stack (PyTorch~2.10.0, CUDA~13.0), and training efficiency strategies (DDP and BF16 mixed precision). Pretraining the largest model (SPERA-Large, 396M) takes approximately 287 GPU hours. Section~\ref{sec:ablation} and Table~\ref{tab:efficiency_profiling} additionally report throughput and peak GPU memory for the attention and Legendre prior variants. Additional experiments and ablations were conducted primarily with SPERA-Small (48M), including the matched objective and corpus controls, and therefore required substantially less compute per run than the largest-model pretraining.
    
\item {\bf Code of ethics}
    \item[] Question: Does the research conducted in the paper conform, in every respect, with the NeurIPS Code of Ethics \url{https://neurips.cc/public/EthicsGuidelines}?
    \item[] Answer: \answerYes{} % Replace by \answerYes{}, \answerNo{}, or \answerNA{}.
    \item[] Justification: We have reviewed the NeurIPS Code of Ethics and confirm that this research conforms to it in all respects.

\item {\bf Broader impacts}
    \item[] Question: Does the paper discuss both potential positive societal impacts and negative societal impacts of the work performed?
    \item[] Answer: \answerYes{} % Replace by \answerYes{}, \answerNo{}, or \answerNA{}.
    \item[] Justification: SPERA can support clinical applications such as sleep staging, abnormal-event detection, and mental disorder screening, as well as assistive BCI applications including motor imagery and imagined speech decoding, particularly in settings with limited labeled data. On the negative side, EEG-based models could be misused for unconsented brain-state inference or surveillance, and premature clinical deployment without additional validation could lead to incorrect diagnostic decisions. Fairness considerations also arise from potential demographic imbalance in the pretraining corpus, as noted in Section~\ref{sec:limitations}. We therefore caution against direct clinical or consumer deployment without further validation, fairness audits, and appropriate human oversight.
    
\item {\bf Safeguards}
    \item[] Question: Does the paper describe safeguards that have been put in place for responsible release of data or models that have a high risk for misuse (e.g., pre-trained language models, image generators, or scraped datasets)?
    \item[] Answer: \answerNA{} % Replace by \answerYes{}, \answerNo{}, or \answerNA{}.
    \item[] Justification: The released model is a representation encoder for EEG signals and does not generate content; it does not pose the high-risk misuse concerns associated with generative language or image models. The pretraining corpus consists of EEG recordings from established public repositories (e.g., OpenNeuro, MOABB, PhysioNet, NSRR, TUH) under their respective data use agreements, with no scraped or unverified content (Appendix~\ref{app:ptds}).

\item {\bf Licenses for existing assets}
    \item[] Question: Are the creators or original owners of assets (e.g., code, data, models), used in the paper, properly credited and are the license and terms of use explicitly mentioned and properly respected?
    \item[] Answer: \answerYes{} % Replace by \answerYes{}, \answerNo{}, or \answerNA{}.
    \item[] Justification: All baseline models, prior EEG foundation models, pretraining datasets, and downstream datasets are properly cited at first mention (Sections~\ref{sec:pretraining} and \ref{sec:eval_protocol}; Appendix~\ref{app:ptds}). Appendix~\ref{app:ptds} (Table~\ref{tab:appendix_b2_datasets}) lists each of the 106 pretraining datasets together with its source platform and license (e.g., CC0, CC BY 4.0, CC BY-NC-SA 4.0, GPL-3.0, and the NEDC, NSRR, Brainclinics, and BDSP data use agreements). The software stack is reported with version numbers in Appendix~\ref{app:computing_resources} (PyTorch 2.10.0, transformers 5.3.0, accelerate 1.13.0, MNE-Python 1.11.0). All assets are used in compliance with their respective licenses and terms of use.

\item {\bf New assets}
    \item[] Question: Are new assets introduced in the paper well documented and is the documentation provided alongside the assets?
    \item[] Answer: \answerYes{} % Replace by \answerYes{}, \answerNo{}, or \answerNA{}.
    \item[] Justification: We release SPERA, a new pretrained EEG foundation model, together with training, evaluation, and preprocessing code. The architecture, training procedure, hyperparameters, and computing environment are documented in Section~\ref{sec:methods}, Section~\ref{sec:pretraining}, and Appendices~\ref{app:computing_resources}--\ref{app:impl}.

\item {\bf Crowdsourcing and research with human subjects}
    \item[] Question: For crowdsourcing experiments and research with human subjects, does the paper include the full text of instructions given to participants and screenshots, if applicable, as well as details about compensation (if any)? 
    \item[] Answer: \answerNA{} % Replace by \answerYes{}, \answerNo{}, or \answerNA{}.
    \item[] Justification: This work does not involve any crowdsourcing or new data collection from human subjects. All EEG data are obtained from existing public repositories under their respective data use agreements (Appendix~\ref{app:ptds}).

\item {\bf Institutional review board (IRB) approvals or equivalent for research with human subjects}
    \item[] Question: Does the paper describe potential risks incurred by study participants, whether such risks were disclosed to the subjects, and whether Institutional Review Board (IRB) approvals (or an equivalent approval/review based on the requirements of your country or institution) were obtained?
    \item[] Answer: \answerNA{} % Replace by \answerYes{}, \answerNo{}, or \answerNA{}.
    \item[] Justification: This work does not involve new research with human subjects. All EEG recordings come from existing public datasets whose original collection was approved by the relevant institutional review boards or equivalent oversight bodies, as documented in their respective source publications (Appendix~\ref{app:ptds}).

\item {\bf Declaration of LLM usage}
    \item[] Question: Does the paper describe the usage of LLMs if it is an important, original, or non-standard component of the core methods in this research? Note that if the LLM is used only for writing, editing, or formatting purposes and does \emph{not} impact the core methodology, scientific rigor, or originality of the research, declaration is not required.
    %this research? 
    \item[] Answer: \answerNA{} % Replace by \answerYes{}, \answerNo{}, or \answerNA{}.
    \item[] Justification: LLMs were not used as a component of the core methodology of this research.

\end{enumerate}

\end{document}